\documentclass[11pt,a4paper]{article}

\usepackage[utf8]{inputenc}
\usepackage[T1]{fontenc}
\usepackage{lmodern}
\usepackage[margin=1in]{geometry}
\usepackage{microtype}
\usepackage{graphicx}
\usepackage{booktabs}
\usepackage{longtable}
\usepackage{array}
\usepackage{needspace}
\usepackage{multirow}
\usepackage{caption}
\usepackage{subcaption}
\usepackage{float}
\usepackage{placeins}
\usepackage{titlesec}
\titlespacing*{\section}{0pt}{1.0ex plus .2ex minus .2ex}{0.6ex plus .1ex}
\titlespacing*{\subsection}{0pt}{0.8ex plus .2ex minus .2ex}{0.4ex plus .1ex}
\titlespacing*{\subsubsection}{0pt}{0.6ex plus .2ex minus .2ex}{0.3ex plus .1ex}
\usepackage{enumitem}
\usepackage{hyperref}
\usepackage[numbers]{natbib}
\setcitestyle{square}
\usepackage{authblk}

\usepackage{xcolor}

\hypersetup{
  colorlinks=true,
  linkcolor=blue,
  citecolor=blue,
  urlcolor=blue,
  pdfauthor={Niraj Gadhe and Vinay Saini and Kirti Bhardwaj and Moulik Jain and Shubhi Sharma},
  pdftitle={Time Series Network Utilization KPI Forecasting Using Advanced AI/ML Models}
}

\title{Time Series Network Utilization KPI Forecasting Using Advanced AI/ML Models}

\author[1]{Niraj Gadhe}
\author[2]{Vinay Saini}
\author[3]{Kirti Bhardwaj}
\author[4]{Moulik Jain}
\author[5]{Shubhi Sharma}

\affil[1,2,3,4,5]{Independent Researcher}

\date{}

\begin{document}

\maketitle

\begin{abstract}
The rapid proliferation of data-intensive applications, cloud infrastructure, and IoT ecosystems has made proactive resource provisioning critical for maintaining optimal network performance. However, network administrators face a constant battle against capacity constraints, where traditional reactive approaches fail to accurately anticipate traffic fluctuations. This inability to foresee demand leads to costly over-provisioning, unexpected downtime, and degraded quality of service---directly impacting operational budgets and business continuity. To achieve efficient capacity planning, accurate forecasting of bandwidth utilization is essential. This study addresses the challenge by evaluating a diverse spectrum of models---including seasonal decomposition, Prophet, Random Forest, XGBoost, Support Vector Regression, and advanced deep learning architectures like bidirectional and Convolutional LSTMs---using a common interface dataset benchmarked across MAPE, NRMSE, and $R^2$ metrics. Ultimately, this research delivers actionable insights into the trade-offs between model accuracy and computational efficiency, empowering engineers, operators, and business owners to select the optimal forecasting model for their specific infrastructure needs.
\end{abstract}


\newpage

\section{Introduction}
\label{sec:introduction}
With the rapid increase in data-intensive applications, cloud infrastructure, and IoT ecosystems~\cite{ref01}, network administrators are challenged to ensure optimal performance, minimizing congestion, and proactively provisioning resources~\cite{ref01}. Accurate forecasting of network resources is therefore essential for efficient capacity planning and maintaining high quality of service~\cite{ref02}.

While predictive techniques can be applied to various network metrics such as traffic, latency, jitter, and packet loss, this paper specifically focuses on forecasting bandwidth utilization---a key metric for efficient network operations~\cite{ref03}. To achieve this, we explore a range of predictive models including Seasonal Decomposition and Prophet (Additive Time Series Model)~\cite{ref04}, tree-based models like Random Forest and XGBoost, as well as supervised regression models like Support Vector Regression (SVR)~\cite{ref05}. Additionally, this research incorporates advanced time series analysis techniques and architecture, utilizing stacked and bidirectional Long Short-Term Memory (LSTM) networks, neural network-based models such as Convolutional Neural Networks (CNN) and Recurrent Neural Networks (RNN), to effectively capture complex temporal dynamics in network performance~\cite{ref01}.

Each algorithm is applied to a common interface utilization dataset and evaluated using key performance metrics such as Mean Absolute Percentage Error (MAPE), Normalized Root Mean Squared Error (NRMSE), and $R^2$ to assess model accuracy and computational efficiency~\cite{ref06}. Ultimately, this research aims to provide actionable insights into the strengths and limitations of different forecasting methods, helping stakeholders such as Technical Engineers, Business Owners and Network Operators select the most effective models for their specific bandwidth forecasting needs.

\clearpage
\section{Challenges}
\label{sec:challenges}
Driven by the accelerated evolution of technology, we are leveraging Artificial Intelligence (AI) within this data-centric environment to transform information into a catalyst for growth, efficiency, and reliability.

However, increasing complexity and scale of AI applications have surfaced several challenges, prompting us to conduct this study. One of the key issues we encountered was the inability to accurately predict bandwidth usage, which led to inefficient capacity planning, underutilized resources in some regions, and congestion or service degradation in other regions. These challenges were compounded by the inconsistent data collected from multiple sources. Missing data due to faulty connections and other underlying problems, noisy data, and outliers further caused another barrier~\cite{ref07}.

Beyond the data itself, the high computational need for advanced AI models, particularly in the training phase, required substantial resources---processing power, storage, and expert involvement~\cite{ref08}. As our interactions and data volume grew, we realized existing infrastructure lacked the stability to handle real-time traffic~\cite{ref09}.

Moreover, many of the models lacked transparency. Real-time processing emerged as another crucial need---we needed models which can adjust according to real traffic and provide accurate prediction over time~\cite{ref10}. Lastly, maintaining this model over time proved challenging due to concept drift. Monitoring and retraining became essential to ensure sustained accuracy~\cite{ref11}. Coupled with the rising cost of running and maintaining this model, it became clear that deeper study was needed. These practical and technical obstacles led us to pursue a more systematic and intelligent approach to time series forecasting using AI~\cite{ref11}.

\clearpage
\section{Methodology and Approach}
\label{sec:methodology}
This section outlines the fundamental components and methodologies integral to developing robust AI/ML models for forecasting network utilization of KPIs~\cite{ref12}. We explore suitable algorithms, model architectures, and problem-framing approaches, including illustrative charts and tables, that lay the groundwork for effective prediction~\cite{ref13}.

\subsection{Candidate Algorithms and Models}
\label{subsec:candidate-algorithms}
The selection of an appropriate algorithm or model architecture is crucial for accurate time series forecasting~\cite{ref12}. The choice often depends on the specific characteristics of the network utilization data, the desired forecast horizon, the complexity of patterns, and the availability of computational resources~\cite{ref01}.

Some of the key categories considered for this research paper are given below:

\begin{figure}[H]
  \centering
  \includegraphics[width=0.93\linewidth]{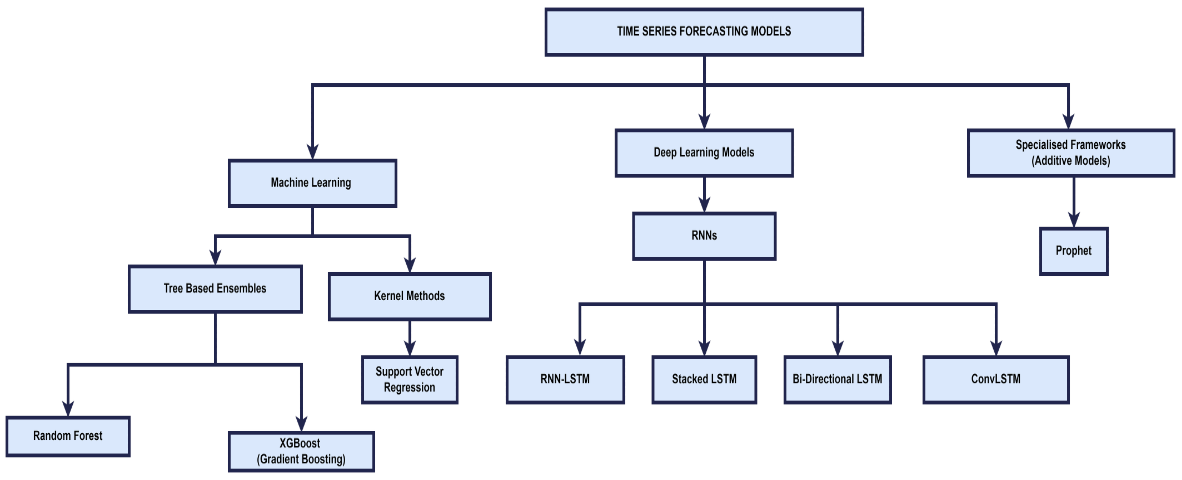}
  \caption{Categorization of Candidate Time Series Forecasting Models/Algorithms.}
  \label{fig:categorization}
\end{figure}

\subsection{Problem Formulation and Analytical Framework}
\label{subsec:problem-formulation}
\Needspace{19\baselineskip}
\noindent Effectively framing the forecasting problem and utilizing appropriate analytical tools and visualizations are foundational to understanding the data and developing accurate models.

\begin{table}[htbp]
  \centering
  \small
  \setlength{\tabcolsep}{4pt}
  \renewcommand{\arraystretch}{1.06}
  \caption{Defining the Forecasting Task: Indicators and Parameters.}
  \label{tab:forecasting-task}
  \begin{tabular}{@{}>{\raggedright\arraybackslash}p{0.26\linewidth}>{\raggedright\arraybackslash}p{0.70\linewidth}@{}}
    \toprule
    \textbf{Parameter} & \textbf{Description} \\
    \midrule
    Identifying Key Performance Indicators (KPIs) for Forecasting &
    The initial step is to explicitly identify the specific network metrics that require prediction. The choice of KPIs will depend on operational objectives, such as capacity planning, anomaly detection, or resource optimization~\cite{ref02}. For our current analysis, we are using Network Bandwidth Utilization~\cite{ref03}.\newline
    \textit{Examples:} Average/Peak Link Bandwidth Utilization (\%), Data Throughput (e.g., Gbps, Mbps), Number of Active Connections/Sessions, Network Latency (ms), Packet Loss or Error Rates (\%) \\
    Establishing the Forecast Horizon &
    This parameter defines the future time period for which predictions are needed. The forecast horizon can range from very short-term (e.g., minutes) to long-term (e.g., days or weeks)~\cite{ref02}. The length of the horizon significantly impacts model choice, complexity, and the potential for accumulating forecast error~\cite{ref14}.\newline
    \textit{Horizons:} Short-term: Next 15 minutes, 1 hour; Medium-term: Next 24 hours, 7 days; Long-term: Next 30 days, 1 quarter \\
    Specifying Data Granularity &
    Data granularity refers to the time resolution of both the input data used for model training and the desired forecast output~\cite{ref15}. Finer granularity can capture more detailed patterns but may also introduce more noise, while coarser granularity might smooth over crucial short-term fluctuations~\cite{ref15}.\newline
    \textit{Granularities:} High-frequency: Per-second, 1-minute intervals; Standard: 5-minute, 15-minute, hourly intervals; Low-frequency: daily or weekly intervals \\
    \bottomrule
  \end{tabular}
\end{table}

\subsection{System and Design Requirements}
\label{subsec:system-requirements}
For this study, the system and design requirements specify the use of Jupyter notebooks as the primary computing environment, with CSV files serving as the main format for data storage and exchange. The implementation relies on Python~3.8 and a diverse set of libraries, including NumPy, Pandas, Matplotlib, TensorFlow, Keras, Scikit-learn, XGBoost, Prophet, and Statsmodels, to facilitate data preprocessing, visualization, and model training. Additionally, the system configuration requires 16~GB of RAM and 128~GB of disk space.

\clearpage
\section{Experiments and Analysis}
\label{sec:experiments}
This section outlines the fundamental components and methodologies integral to developing robust AI/ML models for forecasting network utilization of KPIs. We explore suitable algorithms, model architectures, and problem-framing approaches, including illustrative charts and tables, that lay the groundwork for effective prediction.

\subsection{Data Collection and Pre-processing}
\label{subsec:data-collection}
\subsubsection{Data Collection}
\label{subsubsec:data-collection}
Data is collected from various sources further normalized into a Key Performance Indicator (KPI). For evaluating our results, we have used data from 4 different types of interfaces with the data points being sampled uniformly at a fixed 15-minute resolution.

\subsubsection{Dataset Analysis and Preview}
\label{subsubsec:dataset-preview}
In the dataset, represented in Table~\ref{tab:dataset-preview}, each row captures a snapshot at a specific timestamp, recorded under the \texttt{datetime} field, paired with a Key Performance Indicator (KPI) that describes what is being measured, such as utilization rates. The \texttt{index} field identifies the unique network interface, and the \texttt{value} field contains the actual numeric measurement for the specified time, interface, and KPI. Using this structure, the dataset enables detailed network performance analyses over time, including investigations of bandwidth usage, capacity planning, and traffic flow by device or interface.

Additionally, Table~\ref{tab:interface-types} outlines general functions of network interface types used in the creation of the dataset.

\begin{table}[H]
  \centering
  \caption{Dataset Preview.}
  \label{tab:dataset-preview}
  \begin{tabular}{@{}llll@{}}
    \toprule
    \textbf{Date \& Time} & \textbf{KPI} & \textbf{INDEX} & \textbf{VALUE} \\
    \midrule
    19-06-2019 06:55 & Interface\_Utilization & Interface\_01 & 21582.61 \\
    19-06-2019 07:10 & Interface\_Utilization & Interface\_01 & 21443.13 \\
    19-06-2019 07:25 & Interface\_Utilization & Interface\_01 & 19211.76 \\
    19-06-2019 07:40 & Interface\_Utilization & Interface\_02 & 19785.06 \\
    19-06-2019 07:55 & Interface\_Utilization & Interface\_02 & 20128.95 \\
    19-06-2019 08:10 & Interface\_Utilization & Interface\_03 & 20774.73 \\
    19-06-2019 08:25 & Interface\_Utilization & Interface\_03 & 18825.38 \\
    19-06-2019 08:40 & Interface\_Utilization & Interface\_04 & 18162.39 \\
    19-06-2019 08:55 & Interface\_Utilization & Interface\_04 & 16488.59 \\
    19-06-2019 09:10 & Interface\_Utilization & Interface\_04 & 16912.57 \\
    \bottomrule
  \end{tabular}
\end{table}

\begin{table}[H]
  \centering
  \caption{Interface Type and Function.}
  \label{tab:interface-types}
  \begin{tabular}{@{}llp{0.48\linewidth}@{}}
    \toprule
    \textbf{Interface Label} & \textbf{Interface Type} & \textbf{Function} \\
    \midrule
    Interface\_01 & Edge Peering Interface & Connects internal networks to external networks or peers, facilitating efficient routing and secure data exchange. \\
    Interface\_02 & Caching Interface & Enhances performance by storing frequently accessed data locally, reducing latency and bandwidth usage. \\
    Interface\_03 & Core-to-Edge Interface & Connects core networks to edge networks, enabling high-speed data transfer and processing across network segments. \\
    Interface\_04 & Mobile Network Interface & Supports mobile data transmission, ensuring efficient access to network services for mobile devices. \\
    \bottomrule
  \end{tabular}
\end{table}

\subsubsection{Data Cleaning and Visualization}
\label{subsubsec:data-cleaning}
The preprocessing steps include the removal of null values from data. The DateTime index in our dataset is set as the primary index, and data is resampled to an interval of 15 minutes to standardize the time series~\cite{ref07}. Plotting of interface values is performed to visualize the data, enabling a better understanding of underlying patterns and potential anomalies~\cite{ref16}.

\begin{figure}[H]
  \centering
  \includegraphics[width=0.93\linewidth]{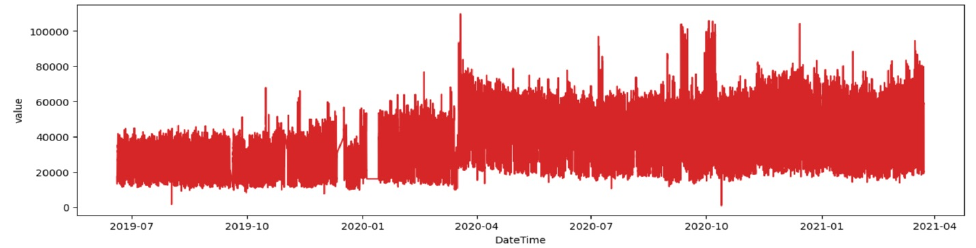}
  \caption{Interface utilization for interface 1.}
  \label{fig:interface1-util}
\end{figure}

\begin{figure}[H]
  \centering
  \includegraphics[width=0.93\linewidth]{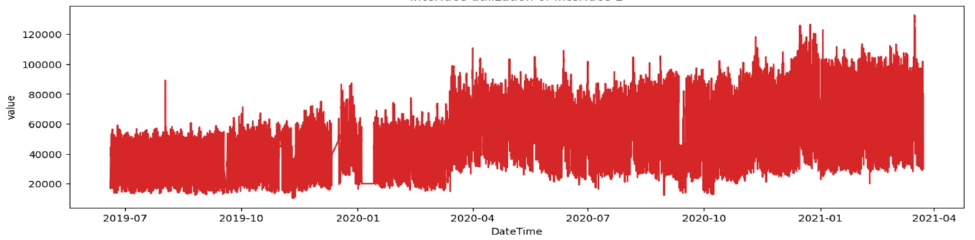}
  \caption{Interface utilization for interface 2.}
  \label{fig:interface2-util}
\end{figure}

\begin{figure}[H]
  \centering
  \includegraphics[width=0.93\linewidth]{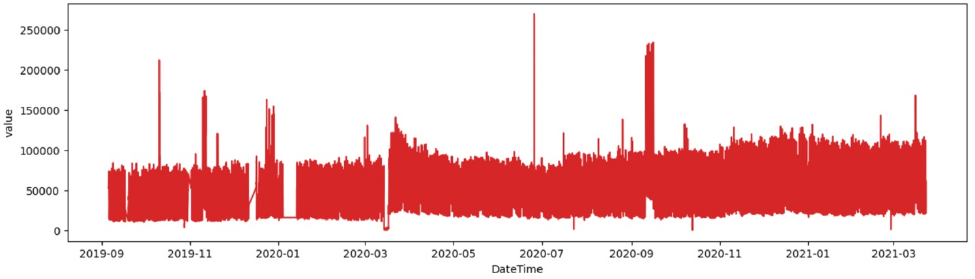}
  \caption{Interface utilization for interface 3.}
  \label{fig:interface3-util}
\end{figure}

\begin{figure}[H]
  \centering
  \includegraphics[width=0.93\linewidth]{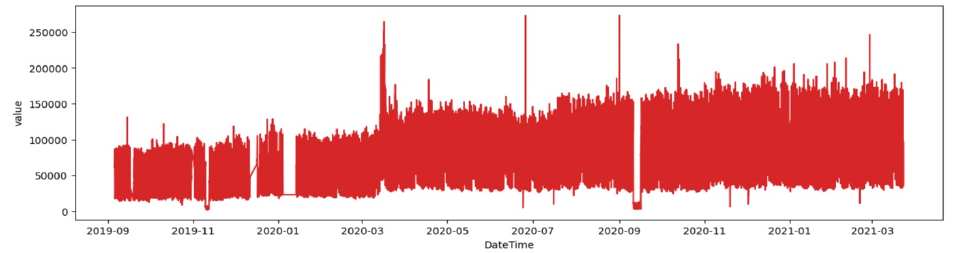}
  \caption{Interface utilization for interface 4.}
  \label{fig:interface4-util}
\end{figure}

\subsubsection{Seasonality, Trend and Periodicity}
\label{subsubsec:seasonality}
In network bandwidth analysis, understanding seasonality, trend, and periodicity is crucial for accurate forecasting and trend identification. Seasonality refers to recurring patterns or fluctuations that occur at regular intervals, often influenced by external factors such as time of day, day of week, or seasonal variations~\cite{ref17}. Trend represents the long-term direction or trajectory of network bandwidth usage, indicating overall growth or decline over time~\cite{ref18}. Periodicity refers to repetitive cycles or oscillations within the data, which may arise from periodic events or systemic patterns~\cite{ref19}.

By analyzing and modeling these components, we can uncover underlying patterns, anticipate future trends, and adapt network management strategies accordingly. Seasonality, trend, and periodicity analysis serve as fundamental pillars in network bandwidth forecasting, enabling proactive resource allocation, capacity planning, and performance optimization~\cite{ref02}.

The dataset used in this experiment is a univariate time series data, focusing on a single KPI over time~\cite{ref20}. Our analysis indicates that the data exhibits no significant seasonal patterns across interfaces. However, we observe a discernible trend within the data, characterized by a slight increase in bandwidth utilization over time. By acknowledging these characteristics, we tailor our forecasting models.

\begin{figure}[H]
  \centering
  \includegraphics[width=0.93\linewidth]{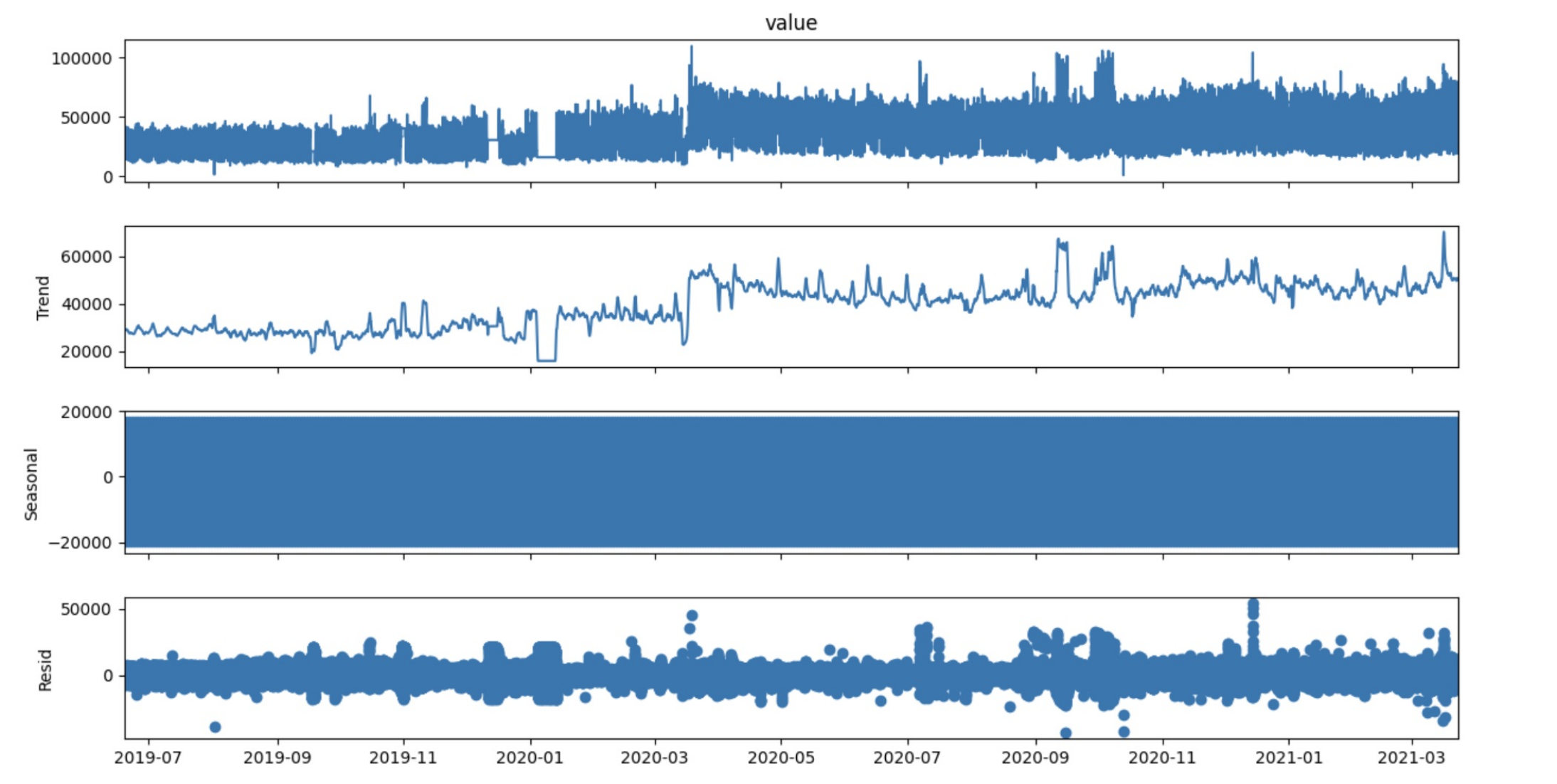}
  \caption{Seasonal decomposition for Interface 1.}
  \label{fig:seasonal-interface1}
\end{figure}

\begin{figure}[H]
  \centering
  \includegraphics[width=0.93\linewidth]{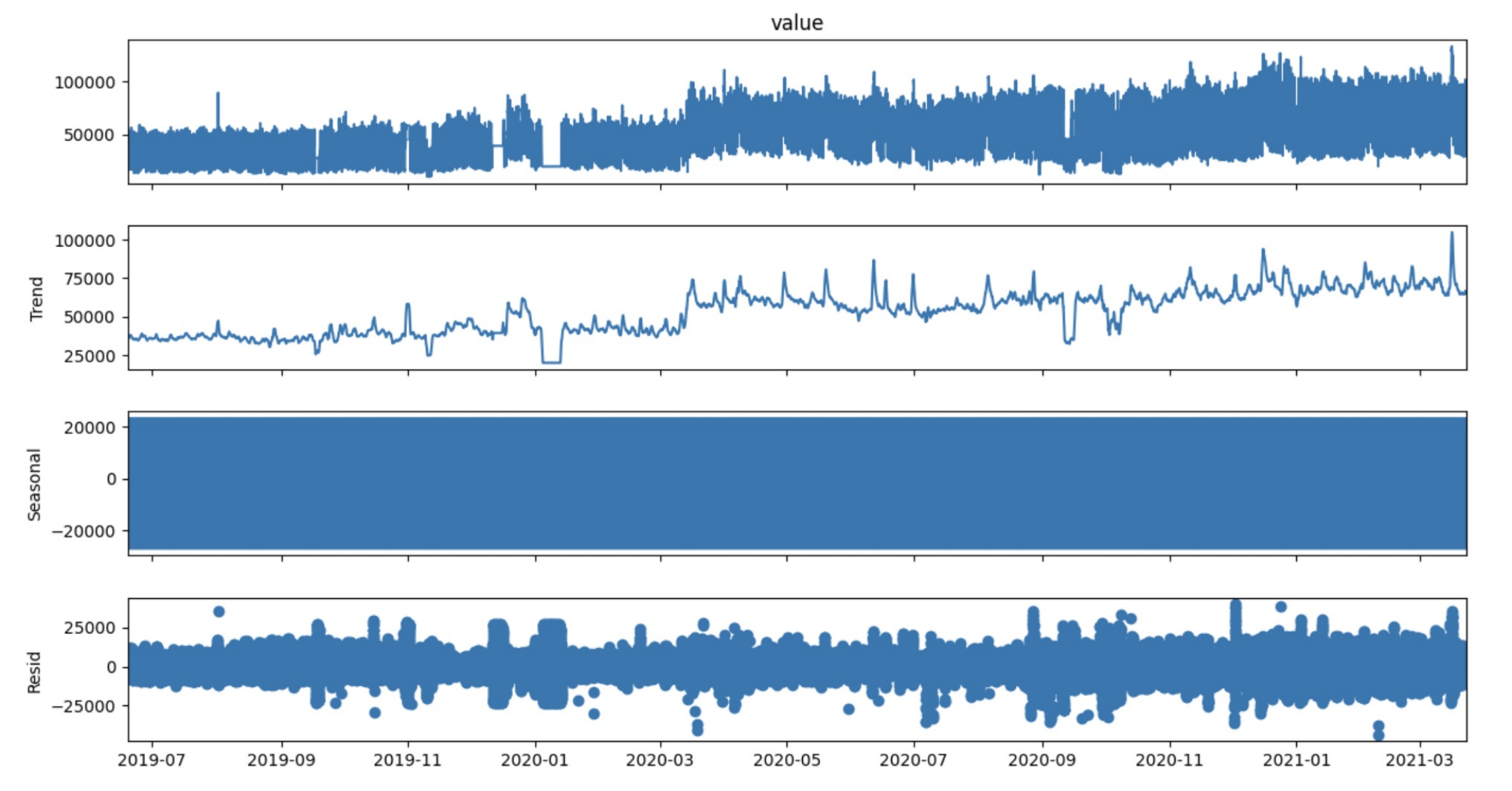}
  \caption{Seasonal decomposition for Interface 2.}
  \label{fig:seasonal-interface2}
\end{figure}

\begin{figure}[H]
  \centering
  \includegraphics[width=0.93\linewidth]{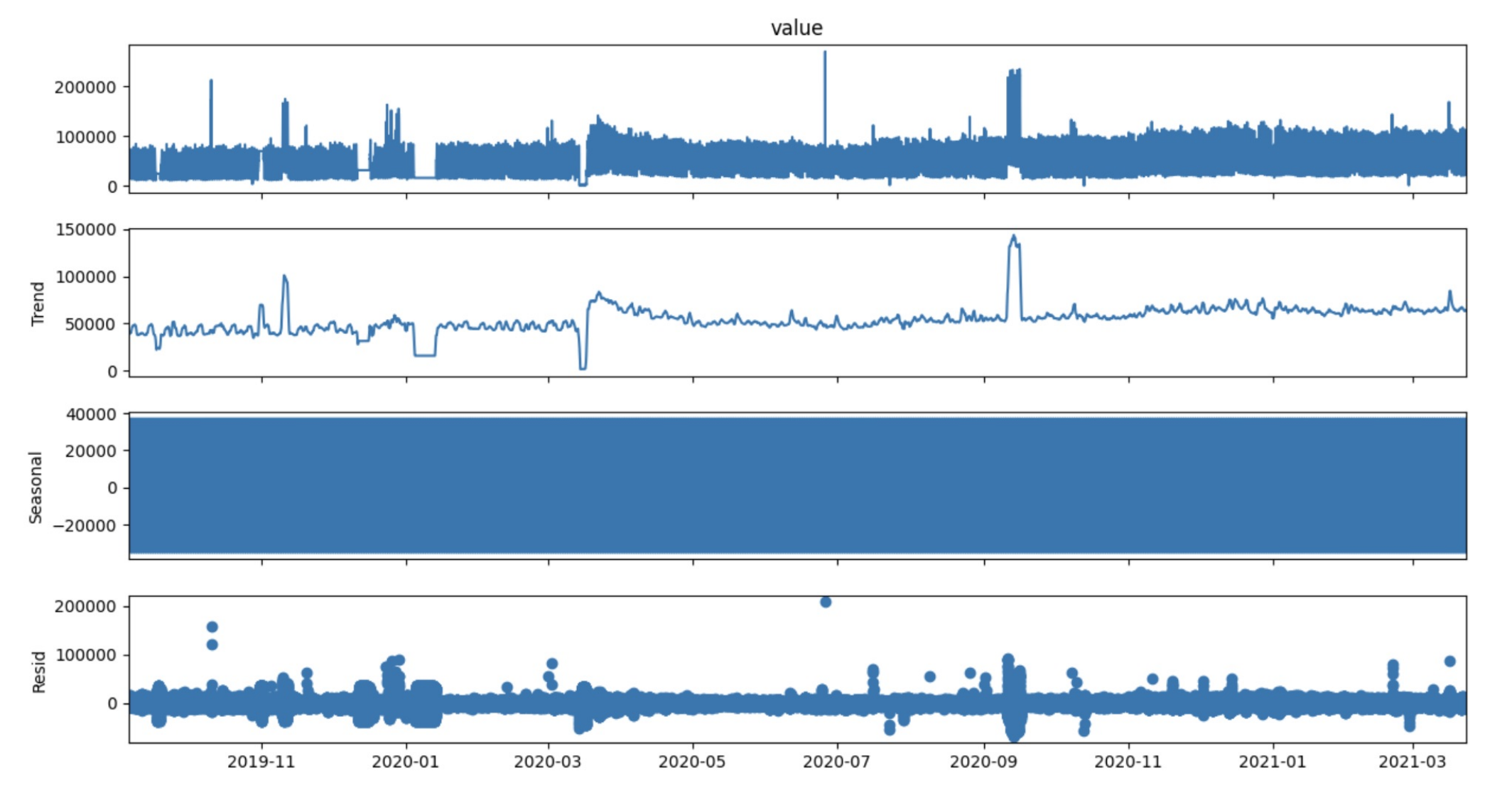}
  \caption{Seasonal decomposition for Interface 3.}
  \label{fig:seasonal-interface3}
\end{figure}

\begin{figure}[H]
  \centering
  \includegraphics[width=0.93\linewidth]{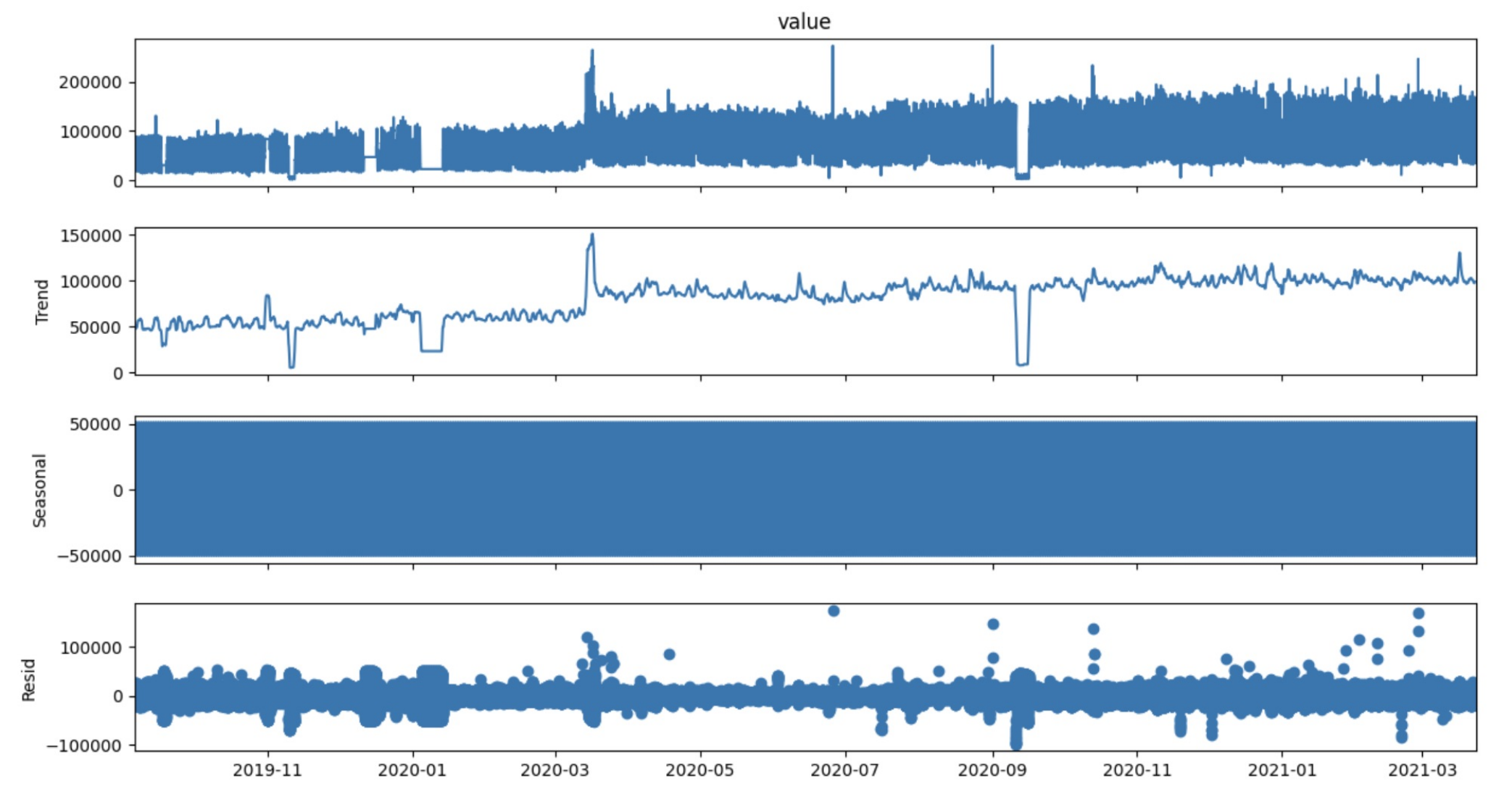}
  \caption{Seasonal decomposition for Interface 4.}
  \label{fig:seasonal-interface4}
\end{figure}

\subsection{Forecasting Algorithms with Hyperparameters}
\label{subsec:hyperparameters}
In network bandwidth forecasting, we employed a diverse array of forecasting algorithms, each tailored to capture specific nuance and pattern in data. In the pursuit of optimizing the performance and predictive accuracy of our forecasting models, careful tuning of hyperparameters plays a crucial role~\cite{ref13}. In this section, we outline the key hyperparameters associated with each forecasting algorithm employed in our analysis.

\subsubsection{Random Forest}
\label{subsubsec:random-forest-hp}
The \texttt{RandomForestRegressor} model was trained without any hyperparameters being explicitly specified or tuned. By default, the model utilizes a set of default hyperparameter values provided by the \texttt{RandomForestRegressor} class in scikit-learn.

\begin{table}[H]
  \centering
  \caption{Hyperparameters for Random Forest Model.}
  \label{tab:rf-hyperparameters}
  \begin{tabular}{@{}ll@{}}
    \toprule
    \textbf{Hyperparameter} & \textbf{Description} \\
    \midrule
    \texttt{n\_estimators} & 100 (Number of trees in the forest) \\
    \texttt{criterion} & \texttt{mse} (Mean Squared Error, used to measure the quality of a split) \\
    \texttt{max\_depth} & \texttt{None} (Nodes are expanded until all leaves are pure or contain fewer samples) \\
    \texttt{min\_samples\_split} & 2 (Minimum number of samples required to split an internal node) \\
    \texttt{min\_samples\_leaf} & 1 (Minimum number of samples required to be at a leaf node) \\
    \texttt{bootstrap} & \texttt{True} (Bootstrap samples are used when building trees) \\
    \texttt{random\_state} & 42 (Ensures reproducibility of results) \\
    \bottomrule
  \end{tabular}
\end{table}

\subsubsection{XGBoost}
\label{subsubsec:xgboost-hp}
The XGBoost model was trained using default hyperparameters without any explicit tuning. XGBoost employs a set of default hyperparameter values, including parameters related to tree boosting, regularization, and optimization algorithms.

\subsubsection{FBProphet}
\label{subsubsec:prophet-hp}
The Prophet model was trained using default hyperparameters without any explicit tuning.

\subsubsection{SVR}
\label{subsubsec:svr-hp}
The parameters used include a regularization parameter ($C$) set to 5, a radial basis function (RBF) kernel with a gamma value of 0.5, and an epsilon value of 0.05. Additionally, the model's degree was set to 5, indicating a polynomial kernel function. Other hyperparameters such as cache size, coefficient 0, maximum iterations, and tolerance were set to their default values provided by the Support Vector Regression implementation in scikit-learn.

\subsubsection{Convolutional LSTM}
\label{subsubsec:convlstm-hp}
This model comprises a stack of layers, including a \texttt{ConvLSTM2D} layer with 64 filters and a $(1, 1)$ kernel size, followed by \texttt{Flatten} and two \texttt{Dense} layers with 32 neurons each. The model is trained using the Adam optimizer and mean squared error loss. This architecture effectively captures spatiotemporal dependencies in the data, facilitating accurate forecasting of network bandwidth utilization.

\begin{figure}[H]
  \centering
  \includegraphics[width=0.93\linewidth]{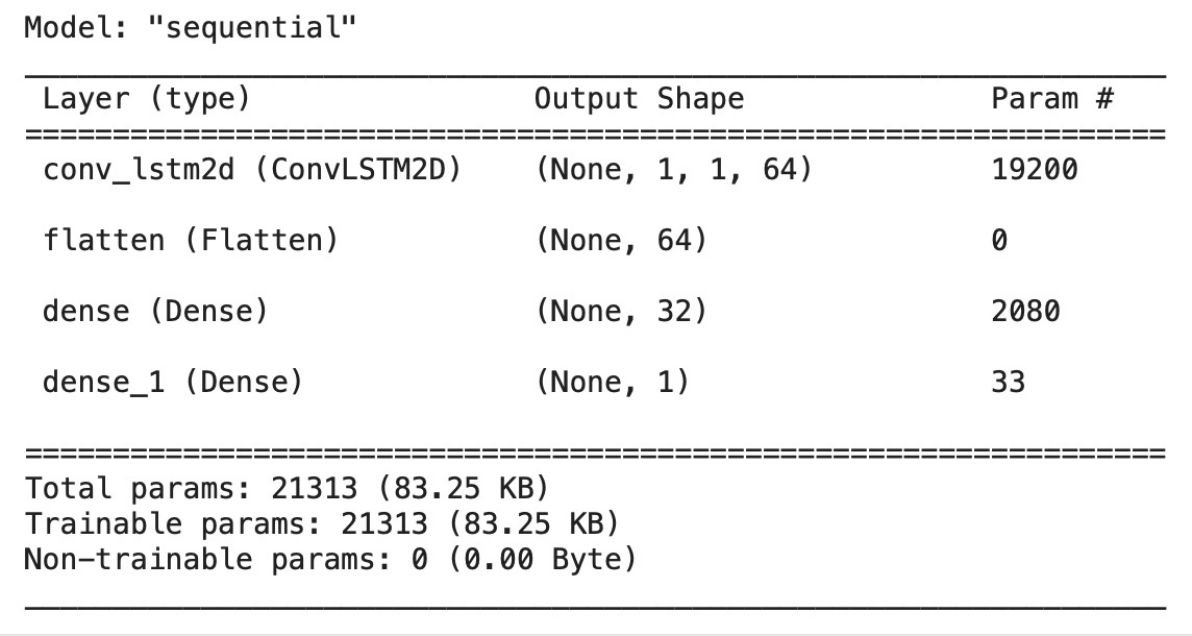}
  \caption{Convolutional LSTM model structure.}
  \label{fig:convlstm-structure}
\end{figure}

\subsubsection{Vanilla/RNN LSTM}
\label{subsubsec:vanilla-lstm-hp}
The model consists of an LSTM layer with 64 units, followed by a Dense layer with 64 units and ReLU activation, and a final Dense output layer with linear activation. The model is compiled with the Adam optimizer and mean absolute error loss function. This architecture effectively captures temporal dependencies in the data, facilitating accurate forecasting of network bandwidth utilization.

\begin{figure}[H]
  \centering
  \includegraphics[width=0.93\linewidth]{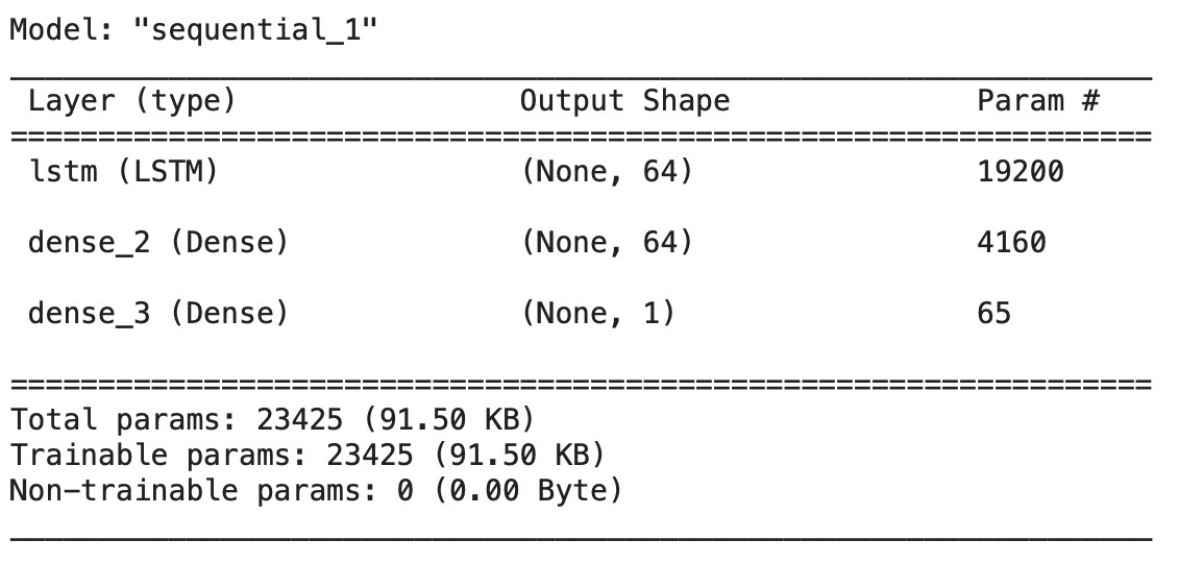}
  \caption{RNN LSTM model structure.}
  \label{fig:rnn-lstm-structure}
\end{figure}

\subsubsection{Stacked LSTM}
\label{subsubsec:stacked-lstm-hp}
This architecture includes two LSTM layers with 64 units each and ReLU activation, configured to return sequences from the first layer. Subsequently, a Dense layer with 64 units and hyperbolic tangent activation is added, followed by a final Dense output layer with linear activation. The model is compiled using the Adam optimizer and mean squared error loss function.

\begin{figure}[H]
  \centering
  \includegraphics[width=0.93\linewidth]{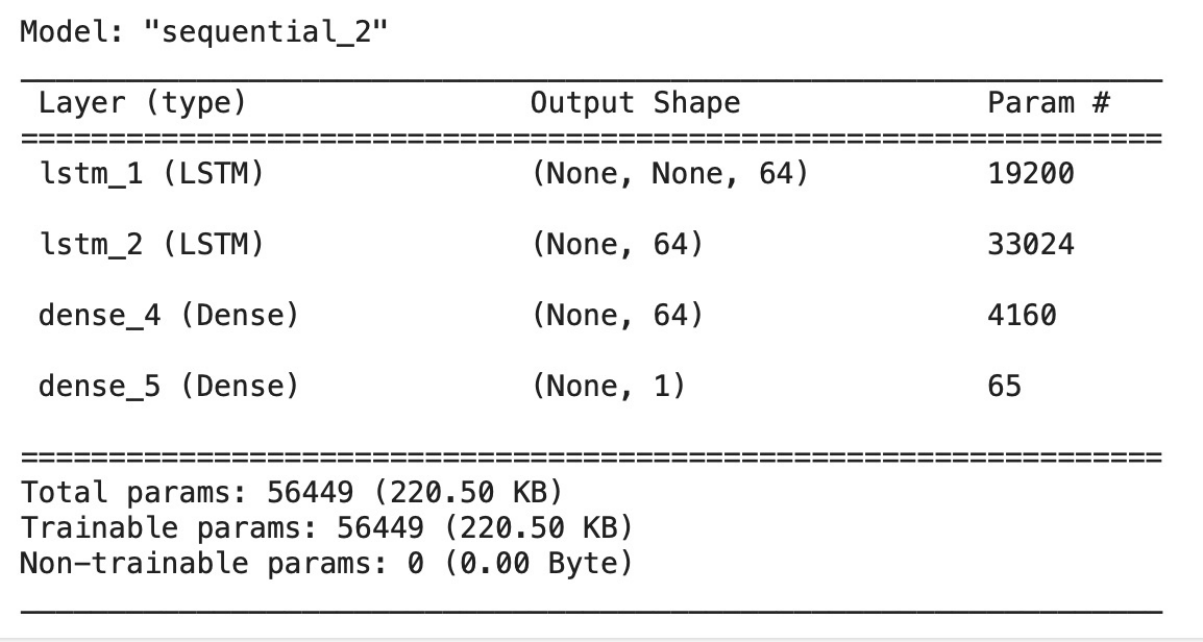}
  \caption{Stacked LSTM model structure.}
  \label{fig:stacked-lstm-structure}
\end{figure}

\subsubsection{Bi-Directional LSTM}
\label{subsubsec:bilstm-hp}
This architecture utilizes a Bidirectional wrapper around an LSTM layer with 64 units and ReLU activation. The model processes input sequences bidirectionally, capturing dependencies from both past and future contexts. A Dense layer with 64 units and hyperbolic tangent activation is added, followed by a final Dense output layer with hyperbolic tangent activation. The model is compiled using the Adam optimizer and mean squared error loss function.

\begin{figure}[H]
  \centering
  \includegraphics[width=0.93\linewidth]{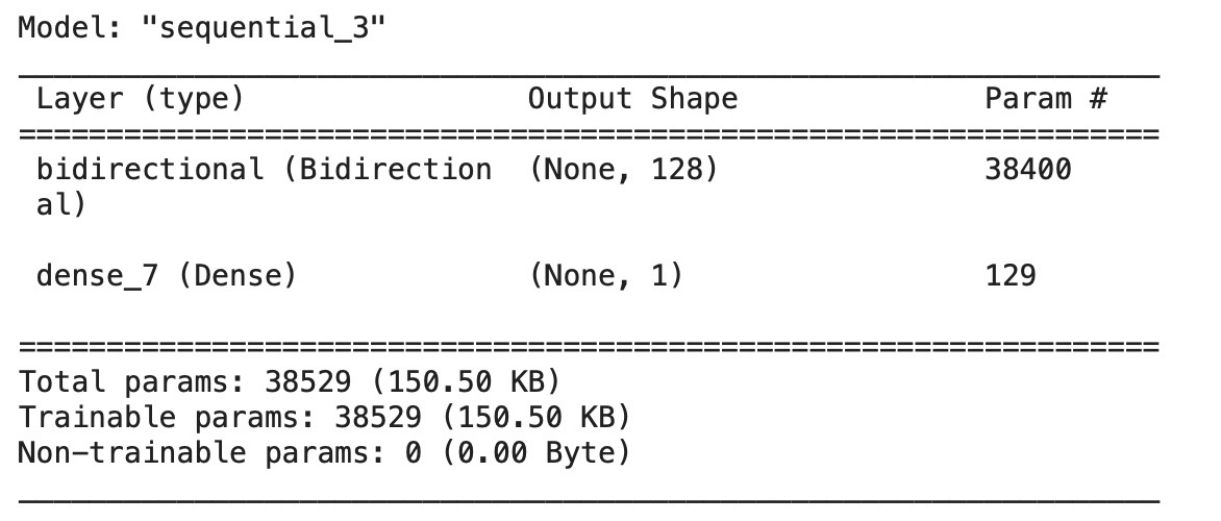}
  \caption{Bi-directional LSTM model structure.}
  \label{fig:bilstm-structure}
\end{figure}

\subsection{Model Training}
\label{subsec:model-training}
\subsubsection{Training}
\label{subsubsec:training}
We used data from four distinct interfaces, employing varying sample sizes ranging from 1000 to 50000 data points from each interface~\cite{ref03}. This diverse dataset selection allowed us to comprehensively capture various aspects of network traffic behavior and performance across multiple interfaces~\cite{ref03}. To ensure robust model evaluation and generalization, we employed a consistent training and testing split ratio of 70:30 across all dataset sizes~\cite{ref21}.

For each algorithm, the input data was structured in a time-series format, consisting of consecutive observations of network bandwidth utilization recorded at discrete time intervals~\cite{ref12}. The dataset was organized into sequences of timestamps $(t_1, t_2, t_3, t_4)$ representing historical observations, with the subsequent timestamp $(t_5)$ reserved for model training~\cite{ref22}. Notably, the sequence size for input varied for each algorithm, accommodating the specific requirements of the modeling techniques employed~\cite{ref23}. During the training of all LSTM models and Support Vector Regression (SVR), a data normalization process was implemented to scale the input features within the range of $(0, 1)$~\cite{ref24}. This normalization was essential to ensure compatibility with the activation functions utilized by these models, namely hyperbolic tangent (tanh) for LSTM models and rectified linear unit (ReLU).

\subsubsection{Performance Metrics}
\label{subsubsec:performance-metrics}
\noindent To assess model accuracy and enable comparative analysis, we used the following key metrics. These helped determine which models performed best across both accuracy and computational efficiency.

\begin{table}[H]
  \centering
  \small
  \setlength{\tabcolsep}{4pt}
  \renewcommand{\arraystretch}{1.06}
  \caption{Key Performance Metrics for Model Evaluation.}
  \label{tab:performance-metrics}
  \begin{tabular}{@{}>{\raggedright\arraybackslash}p{0.26\linewidth}>{\raggedright\arraybackslash}p{0.70\linewidth}@{}}
    \toprule
    \textbf{Metric} & \textbf{Description} \\
    \midrule
    Mean Absolute Percentage Error (MAPE) &
    Calculates the average percentage error between predicted and actual values, offering insight into the relative accuracy of predictions across different scales~\cite{ref25}. \\
    Normalized Root Mean Squared Error (NRMSE) &
    Normalizes RMSE by the range of the target variable, offering a scale-independent accuracy metric~\cite{ref26}. \\
    Training Time &
    The time required to train a model on the dataset, reflecting its computational cost and scalability~\cite{ref27}. \\
    Prediction Time &
    The time taken to generate forecasts on unseen data, indicating the model's efficiency during inference~\cite{ref09}. \\
    $R^2$ Metric &
    Indicates the proportion of variance in the dependent variable that is predictable from the independent variables, assessing the goodness of fit for a model~\cite{ref28}. \\
    \bottomrule
  \end{tabular}
\end{table}

To evaluate model performance, we considered both absolute values and relative comparisons across these metrics~\cite{ref10}. Lower values of MAPE and NRMSE indicate better predictive accuracy, while shorter training and prediction times reflect higher computational efficiency~\cite{ref29}. For MAPE and NRMSE, the following thresholds were used to interpret model accuracy~\cite{ref30}:
\begin{itemize}[noitemsep]
  \item $<10\%$: High accuracy
  \item $10$--$20\%$: Moderate accuracy
  \item $20$--$30\%$: Acceptable
  \item $>30\%$: Unacceptable
\end{itemize}

\clearpage
\section{Evaluation}
\label{sec:evaluation}
Each model was evaluated using the following metrics to identify the most accurate and efficient model for bandwidth utilization forecasting. The results of each are given below.

\subsection{Random Forest}
\label{subsec:eval-rf}
A reliable traditional model with consistent results.

\begin{itemize}[noitemsep]
  \item MAPE was consistently $< 5\%$, confirming high accuracy.
  \item NRMSE was $\leq 22\%$ for most interfaces.
  \item Training time was constant, maxing at 44 seconds per run.
  \item $R^2$ remained above 95\% across interfaces, with Interface~3 highest and Interface~4 most variable.
  \item Prediction time remained very low at small sample sizes ($\sim$0.01--0.06~s) and increased with larger sizes, reaching up to $\sim$0.57~s.
\end{itemize}

\begin{figure}[H]
  \centering
  \includegraphics[width=0.93\linewidth]{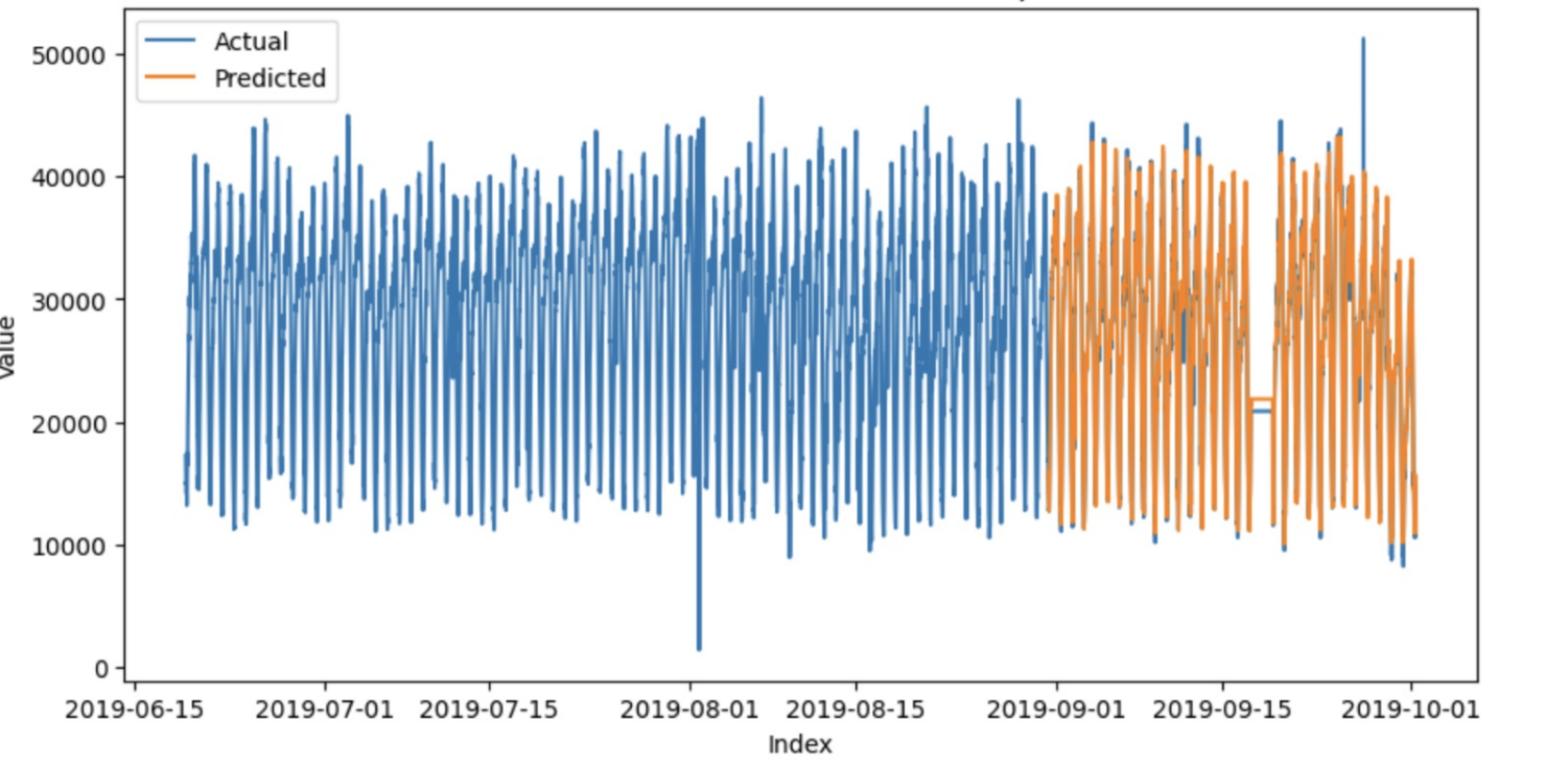}
  \caption{Random Forest: actual vs.\ predicted values.}
  \label{fig:rf-actual-predicted}
\end{figure}

\begin{figure}[H]
  \centering
  \includegraphics[width=0.93\linewidth]{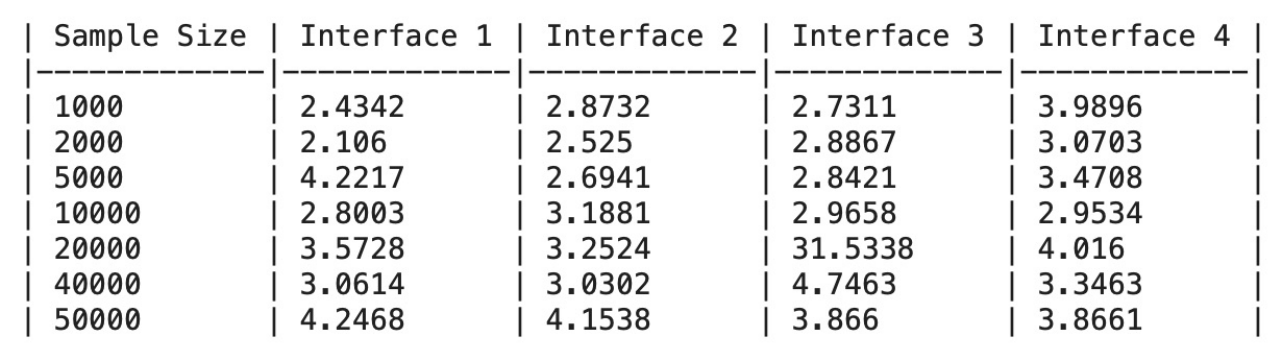}
  \caption{Comparison of MAPE on all interfaces for Random Forest.}
  \label{fig:rf-mape}
\end{figure}

\begin{figure}[H]
  \centering
  \includegraphics[width=0.93\linewidth]{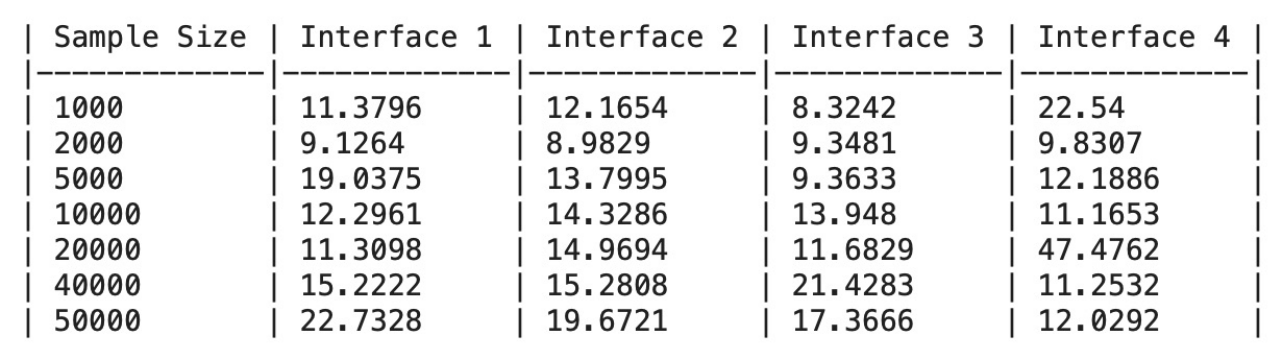}
  \caption{Comparison of NRMSE on all interfaces for Random Forest.}
  \label{fig:rf-nrmse}
\end{figure}

\begin{figure}[H]
  \centering
  \includegraphics[width=0.93\linewidth]{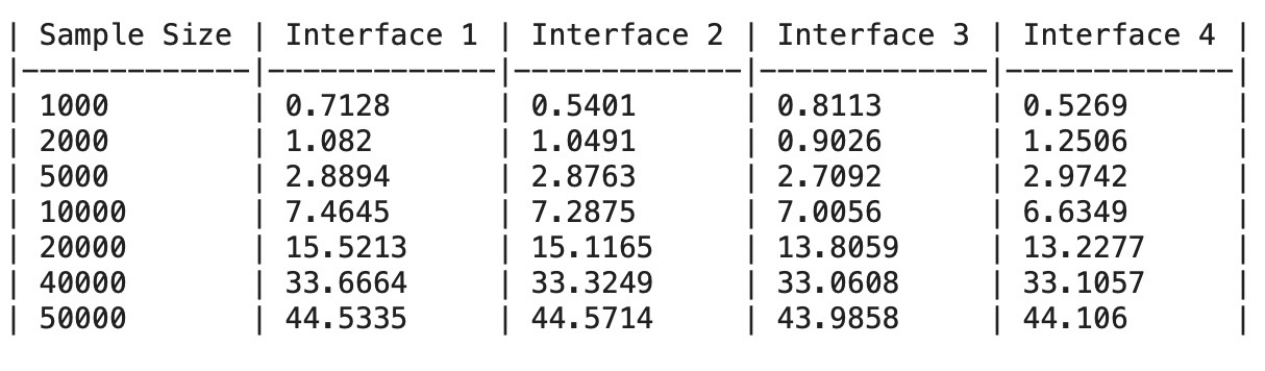}
  \caption{Comparison of training time on all interfaces for Random Forest.}
  \label{fig:rf-training-time}
\end{figure}

\begin{figure}[H]
  \centering
  \includegraphics[width=0.93\linewidth]{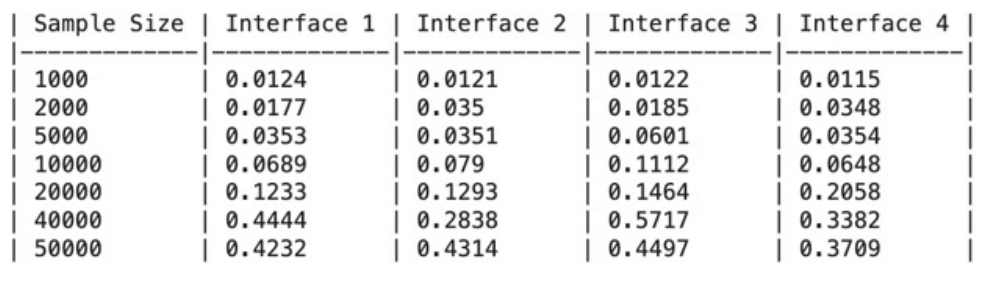}
  \caption{Comparison of prediction time on all interfaces for Random Forest.}
  \label{fig:rf-prediction-time}
\end{figure}

\begin{figure}[H]
  \centering
  \includegraphics[width=0.93\linewidth]{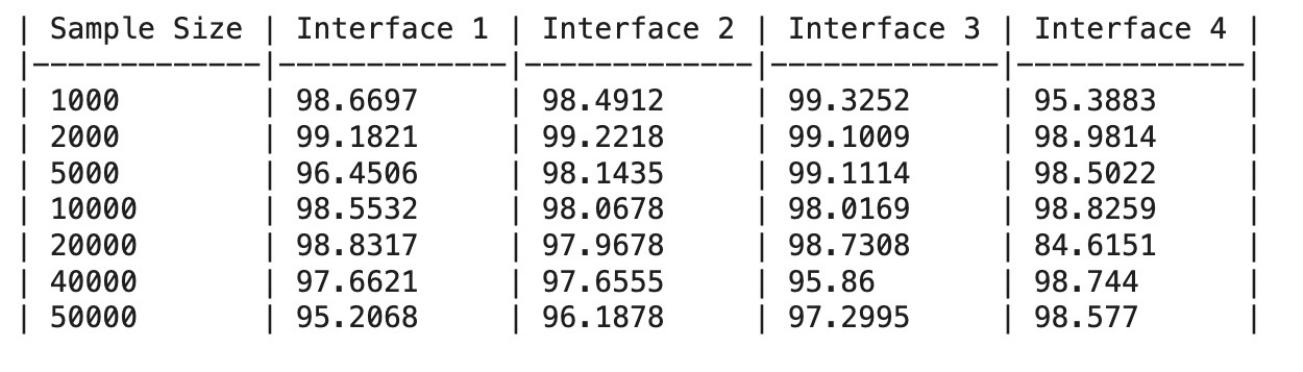}
  \caption{Comparison of $R^2$ metric on all interfaces for Random Forest.}
  \label{fig:rf-r2}
\end{figure}

\subsection{XGBoost}
\label{subsec:eval-xgb}
High-performing and extremely fast, ideal for real-time scenarios.

\begin{itemize}[noitemsep]
  \item MAPE was $< 5\%$, and NRMSE ranged from 5--27\%.
  \item Training time was $< 1$ second, highlighting scalability and speed.
  \item NRMSE ranged from $\sim$8.78\% to $\sim$45.73\%.
  \item Prediction time for XGBoost stayed minimal at small sample sizes ($\sim$0.003--0.009~s) and grew moderately with larger datasets.
  \item $R^2$ remained mostly high ($\sim$94--99\%), with Interface~3 performing strongest overall.
\end{itemize}

\begin{figure}[H]
  \centering
  \includegraphics[width=0.93\linewidth]{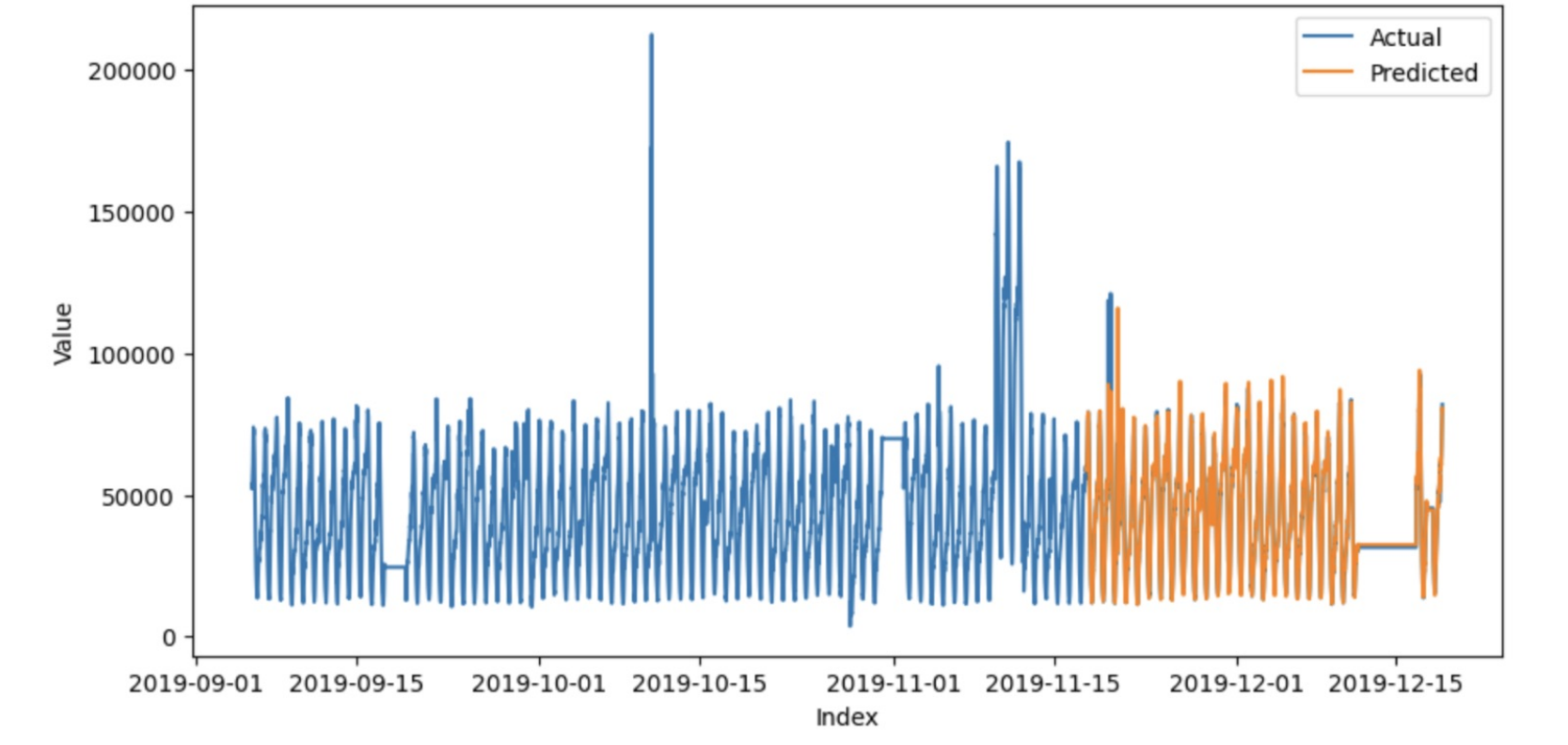}
  \caption{XGBoost: actual vs.\ predicted values for different sample sizes.}
  \label{fig:xgb-actual-predicted}
\end{figure}

\begin{figure}[H]
  \centering
  \includegraphics[width=0.93\linewidth]{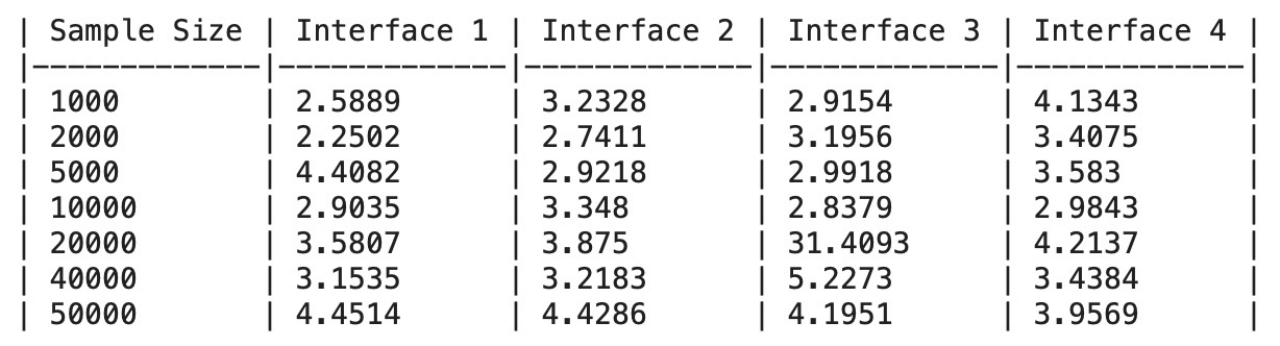}
  \caption{Comparison of MAPE on all interfaces for XGBoost.}
  \label{fig:xgb-mape}
\end{figure}

\begin{figure}[H]
  \centering
  \includegraphics[width=0.93\linewidth]{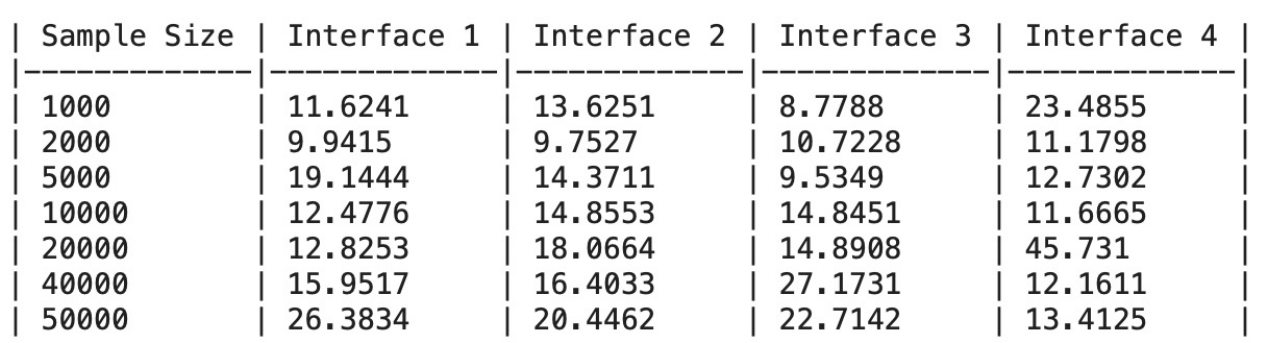}
  \caption{Comparison of NRMSE on all interfaces for XGBoost.}
  \label{fig:xgb-nrmse}
\end{figure}

\begin{figure}[H]
  \centering
  \includegraphics[width=0.93\linewidth]{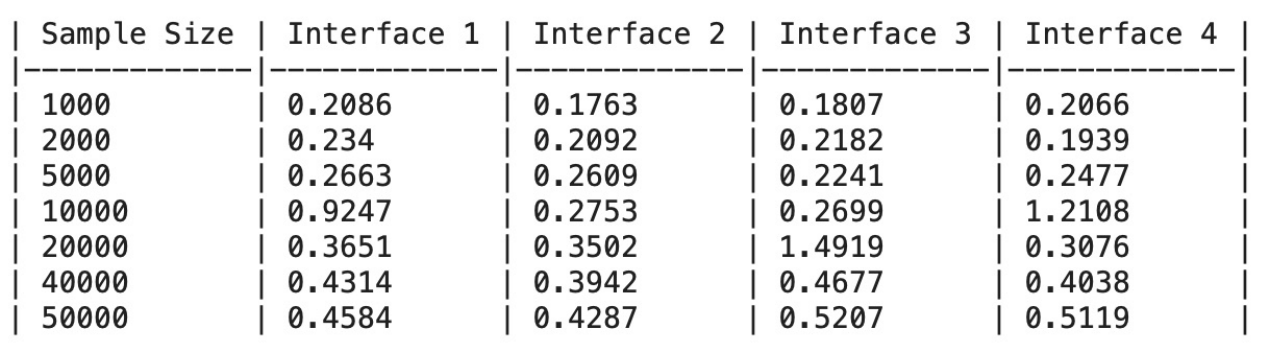}
  \caption{Comparison of training time on all interfaces for XGBoost.}
  \label{fig:xgb-training-time}
\end{figure}

\begin{figure}[H]
  \centering
  \includegraphics[width=0.93\linewidth]{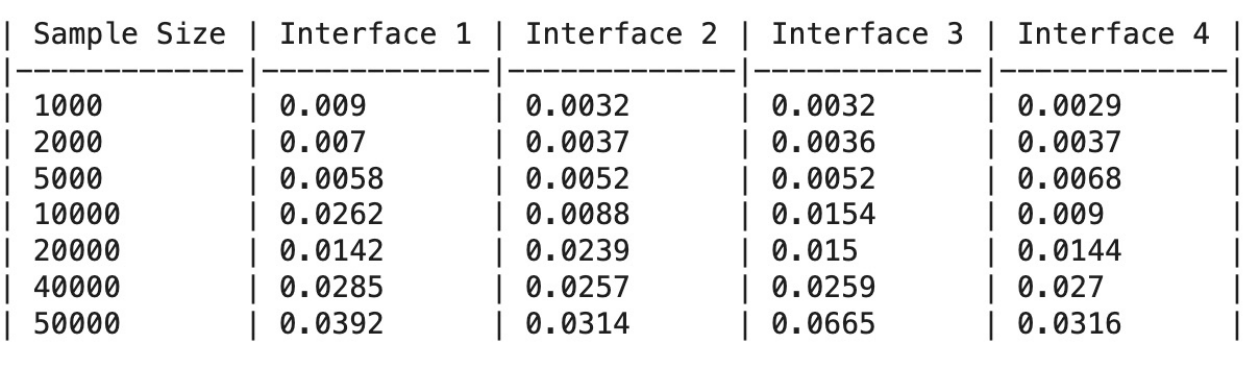}
  \caption{Comparison of prediction time on all interfaces for XGBoost.}
  \label{fig:xgb-prediction-time}
\end{figure}

\begin{figure}[H]
  \centering
  \includegraphics[width=0.93\linewidth]{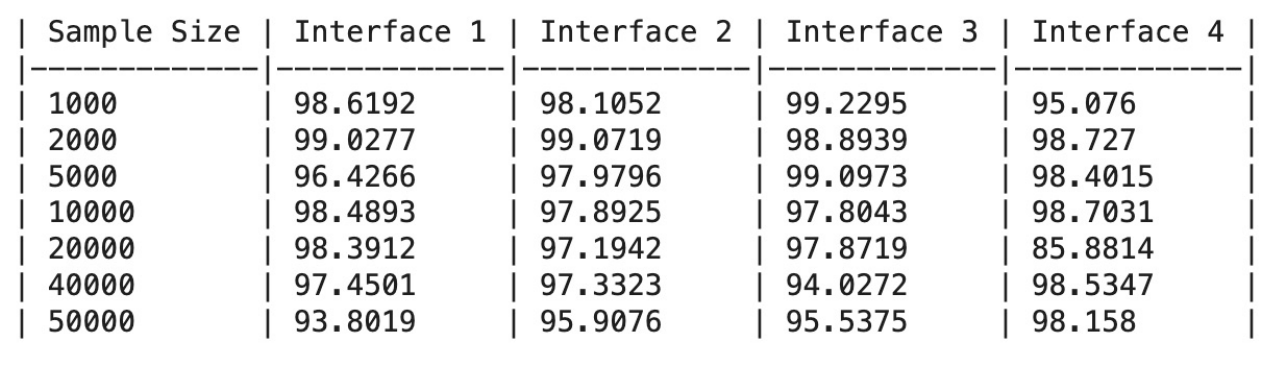}
  \caption{Comparison of $R^2$ on all interfaces for XGBoost.}
  \label{fig:xgb-r2}
\end{figure}

\subsection{Prophet}
\label{subsec:eval-prophet}
Underperformed in current configuration; needs tuning.

\begin{itemize}[noitemsep]
  \item MAPE and NRMSE were mostly $> 20\%$, indicating poor accuracy.
  \item Training time was fast for small data, but averaged $\sim$35 seconds for larger datasets.
  \item NRMSE showed high variability ($\sim$0.03--199.88\%), with Interface~3 performing best overall at most sample sizes, while Interfaces~1 and~2 had occasional large spikes.
  \item $R^2$ varied widely across interfaces, ranging from $-331.68\%$ to 94.24\%, with Interface~4 generally performing best and Interface~2 showing the most negative values at larger sample sizes.
  \item Prediction time for Prophet increased with sample size, ranging from $\sim$0.16~s to $\sim$9.62~s.
  \item Requires hyperparameter tuning for better results.
\end{itemize}

\begin{figure}[H]
  \centering
  \includegraphics[width=0.93\linewidth]{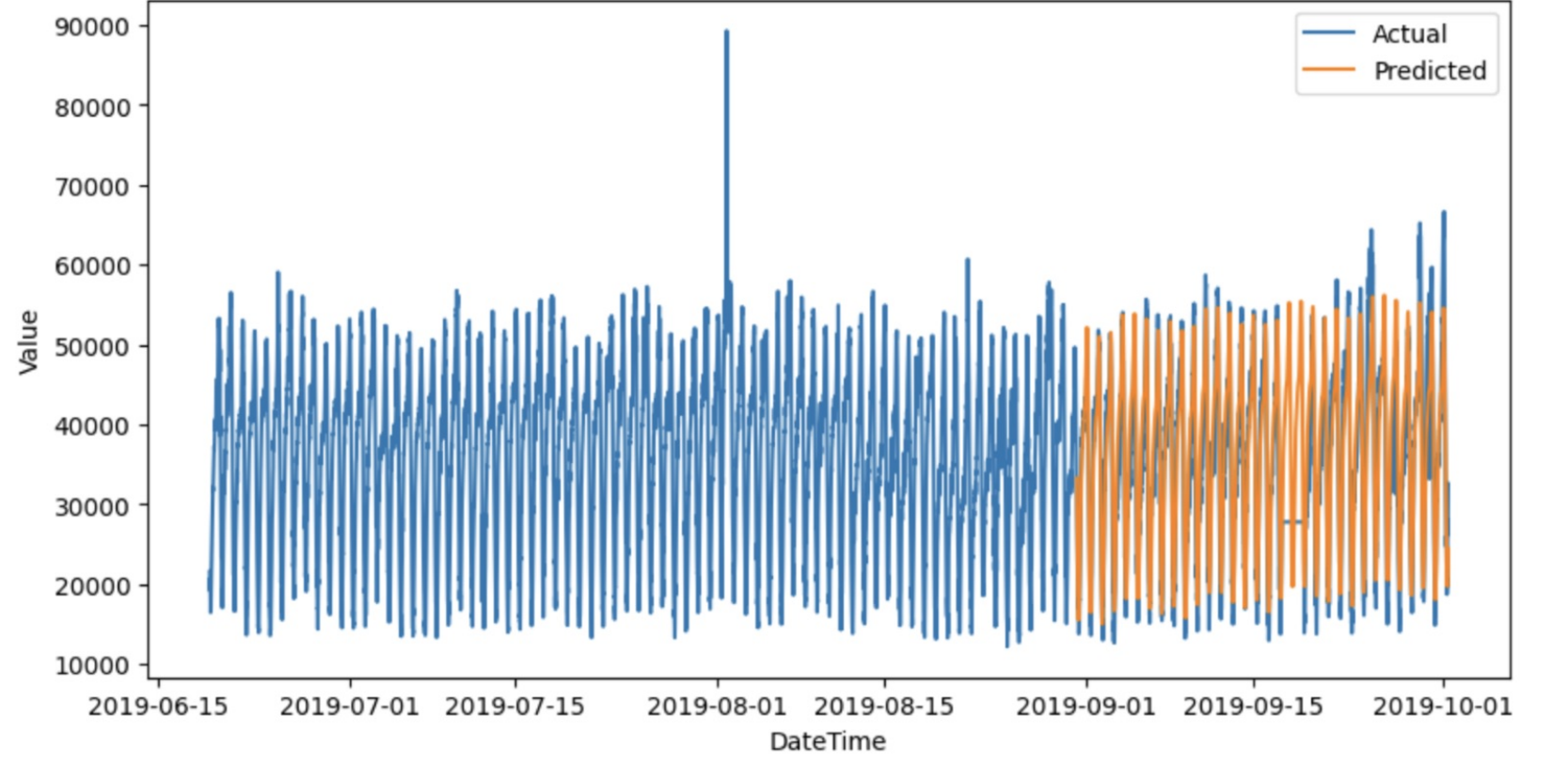}
  \caption{Prophet: actual vs.\ predicted values.}
  \label{fig:prophet-actual-predicted}
\end{figure}

\begin{figure}[H]
  \centering
  \includegraphics[width=0.93\linewidth]{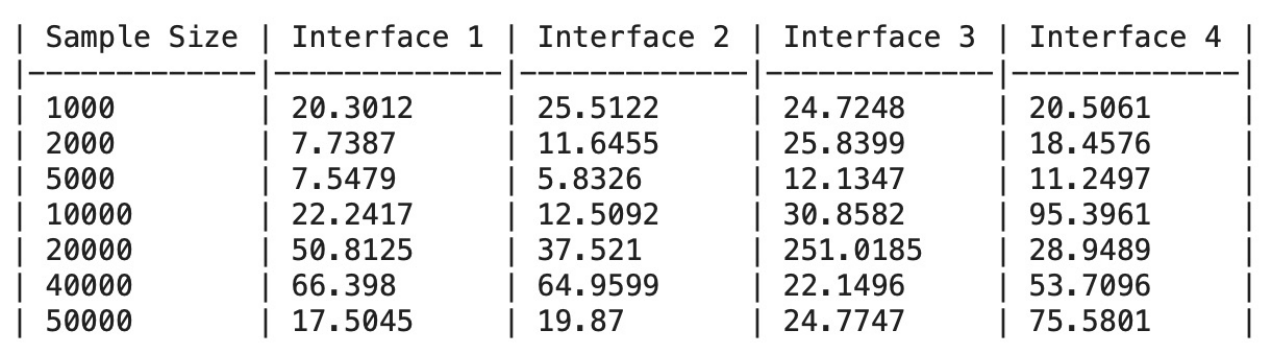}
  \caption{Comparison of MAPE on all interfaces for Prophet.}
  \label{fig:prophet-mape}
\end{figure}

\begin{figure}[H]
  \centering
  \includegraphics[width=0.93\linewidth]{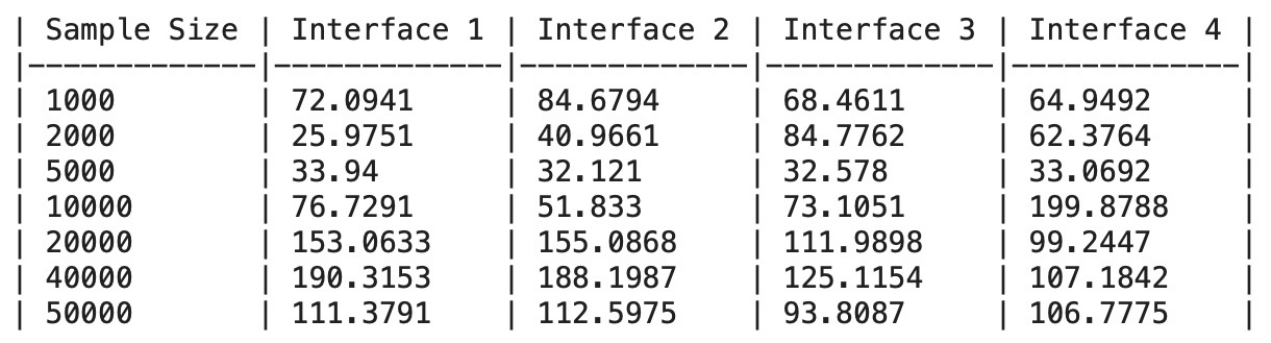}
  \caption{Comparison of NRMSE on all interfaces for Prophet.}
  \label{fig:prophet-nrmse}
\end{figure}

\begin{figure}[H]
  \centering
  \includegraphics[width=0.93\linewidth]{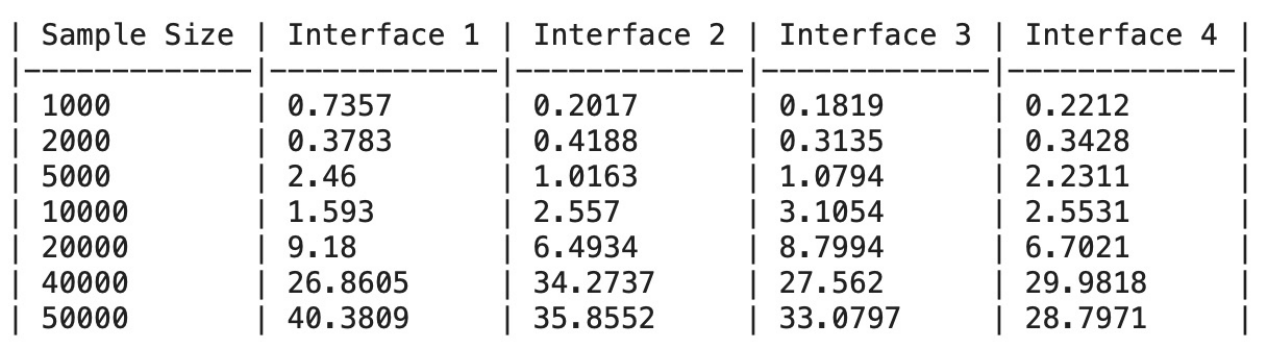}
  \caption{Comparison of training time on all interfaces for Prophet.}
  \label{fig:prophet-training-time}
\end{figure}

\begin{figure}[H]
  \centering
  \includegraphics[width=0.93\linewidth]{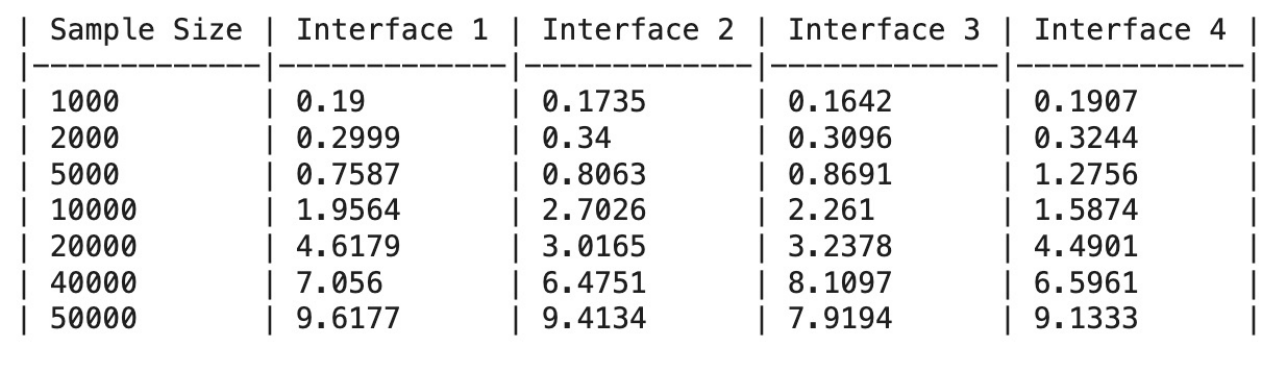}
  \caption{Comparison of prediction time on all interfaces for Prophet.}
  \label{fig:prophet-prediction-time}
\end{figure}

\begin{figure}[H]
  \centering
  \includegraphics[width=0.93\linewidth]{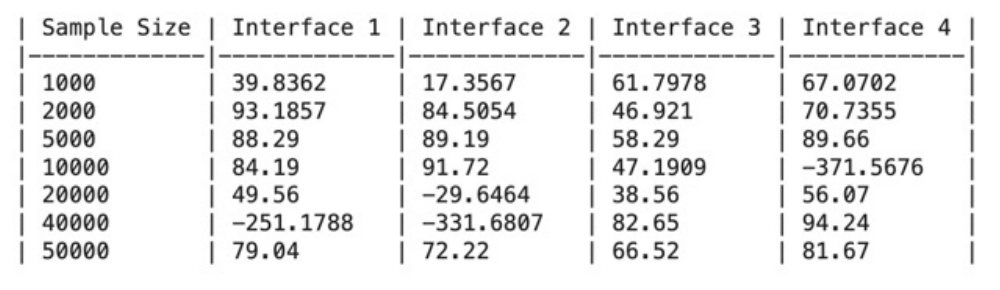}
  \caption{Comparison of $R^2$ metric on all interfaces for Prophet.}
  \label{fig:prophet-r2}
\end{figure}

\subsection{Support Vector Regression (SVR)}
\label{subsec:eval-svr}
Strong accuracy with fast training, suitable for scalable applications.

\begin{itemize}[noitemsep]
  \item MAPE was $< 10\%$ on most interfaces.
  \item NRMSE ranged from 9\% to 28\%, within acceptable bounds.
  \item Training time was $< 3$ seconds, stable across all inputs.
  \item $R^2$ for SVR stayed mostly above 94\%, with Interfaces~1--3 consistently strong ($\sim$94--99\%) and Interface~4 showing high variability, dropping as low as 38.40\% at larger sample sizes.
  \item Prediction time for SVR remained very low at small sample sizes ($\sim$0.002--0.013~s) and increased with larger sizes, reaching up to $\sim$1.01~s, with Interface~2 peaking highest.
\end{itemize}

\begin{figure}[H]
  \centering
  \includegraphics[width=0.93\linewidth]{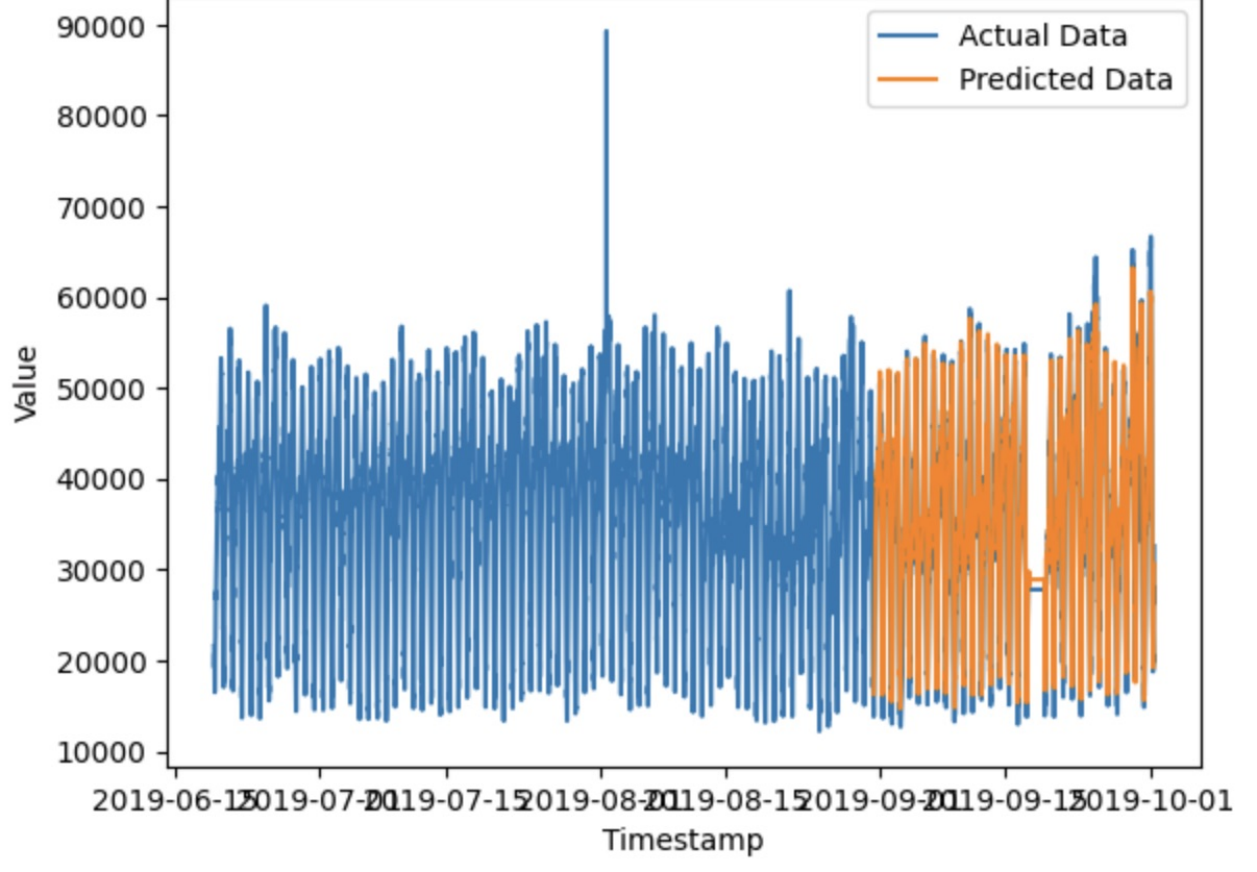}
  \caption{SVR: actual vs.\ predicted values.}
  \label{fig:svr-actual-predicted}
\end{figure}

\begin{figure}[H]
  \centering
  \includegraphics[width=0.93\linewidth]{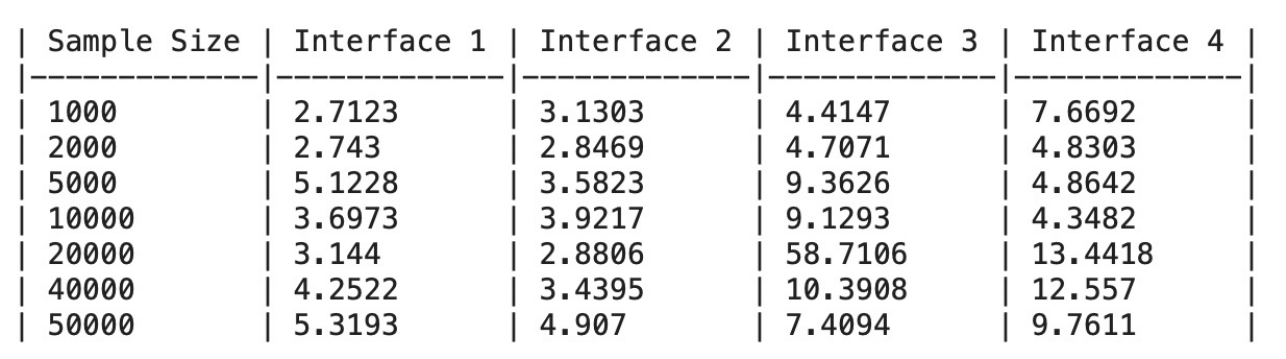}
  \caption{MAPE comparison across interfaces for SVR.}
  \label{fig:svr-mape}
\end{figure}

\begin{figure}[H]
  \centering
  \includegraphics[width=0.93\linewidth]{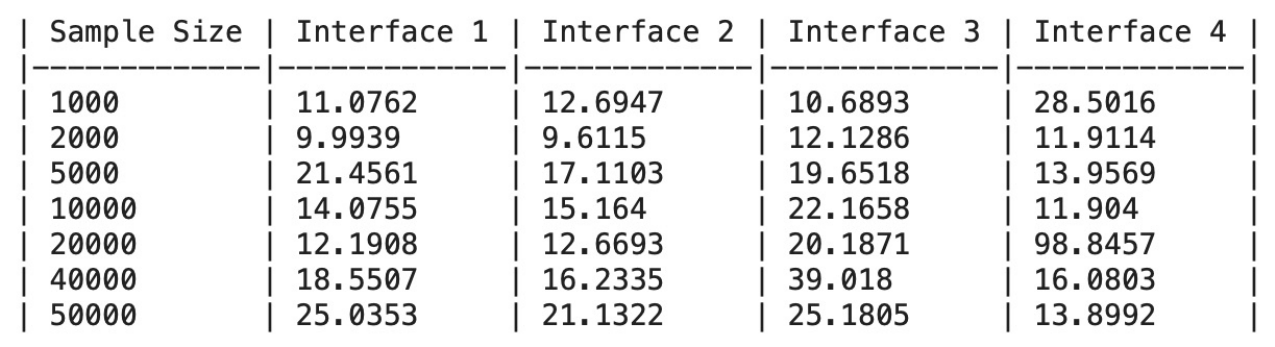}
  \caption{NRMSE comparison across interfaces for SVR.}
  \label{fig:svr-nrmse}
\end{figure}

\begin{figure}[H]
  \centering
  \includegraphics[width=0.93\linewidth]{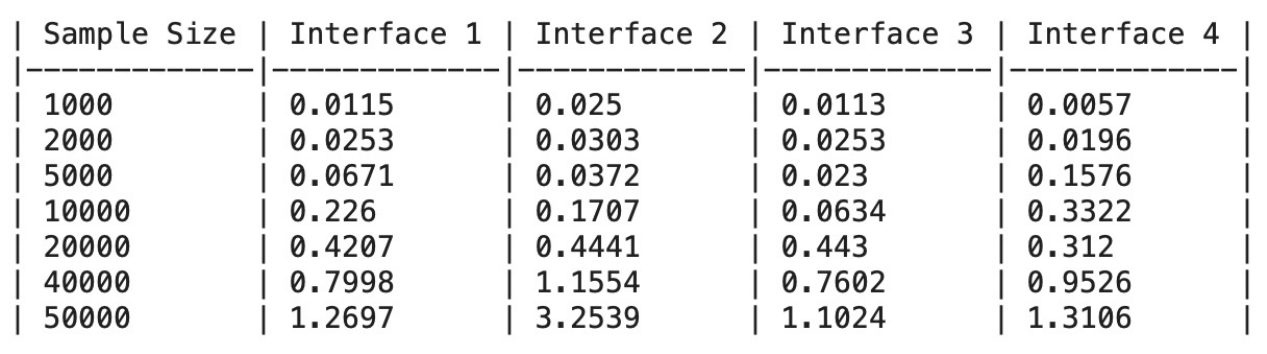}
  \caption{Comparison of training time across interfaces for SVR.}
  \label{fig:svr-training-time}
\end{figure}

\begin{figure}[H]
  \centering
  \includegraphics[width=0.93\linewidth]{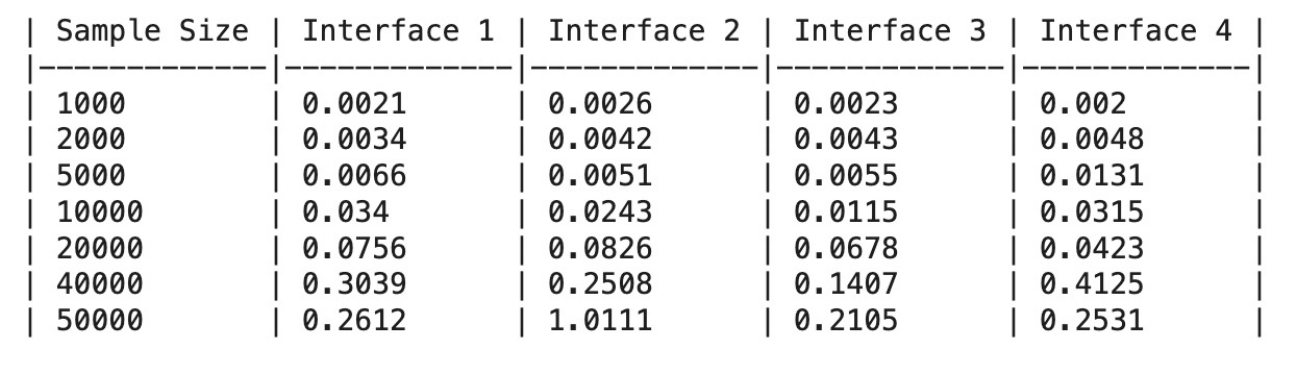}
  \caption{Comparison of prediction time on all interfaces for SVR.}
  \label{fig:svr-prediction-time}
\end{figure}

\begin{figure}[H]
  \centering
  \includegraphics[width=0.93\linewidth]{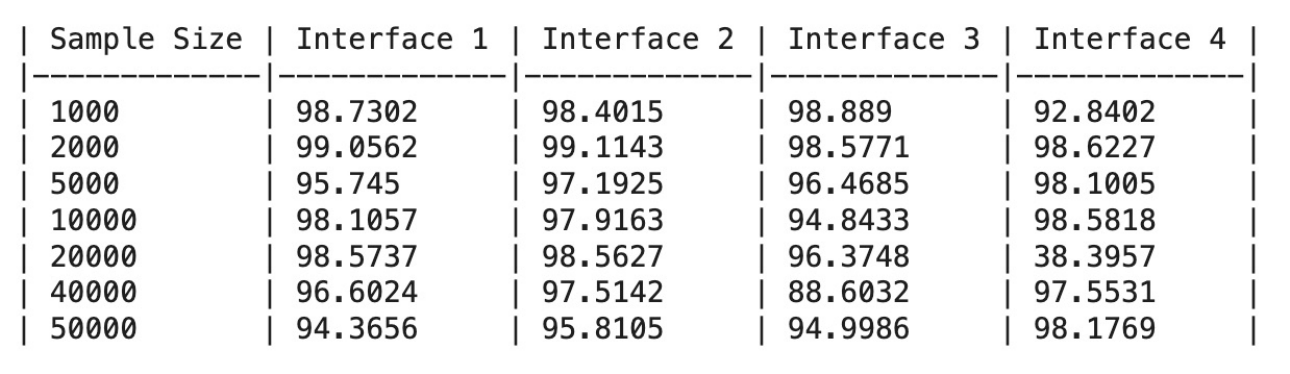}
  \caption{Comparison of $R^2$ metric across interfaces for SVR.}
  \label{fig:svr-r2}
\end{figure}

\subsection{Convolutional LSTM}
\label{subsec:eval-convlstm}
This model demonstrated strong predictive performance across all interfaces.

\begin{itemize}[noitemsep]
  \item MAPE remained consistently below 5\%, indicating high accuracy.
  \item NRMSE ranged between 10--25\%, showing acceptable variability.
  \item Training time increased with sample size but remained under 1 minute, ensuring practical usability.
  \item Prediction time remained under 1.32~s across all interfaces, with Interface~3 consistently fastest ($\sim$0.12--0.76~s).
  \item $R^2$ metric remained above 94\% across all interfaces.
\end{itemize}

\begin{figure}[H]
  \centering
  \includegraphics[width=0.93\linewidth]{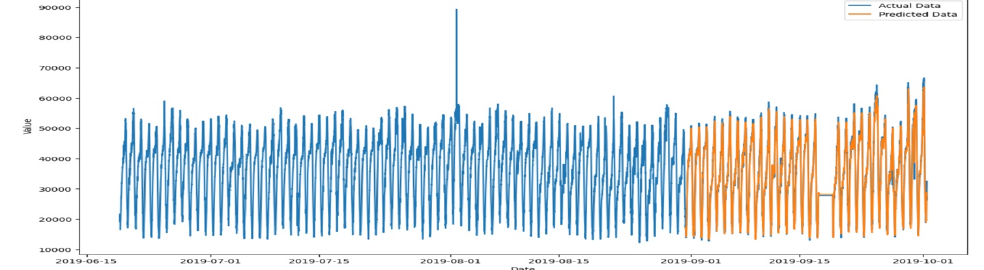}
  \caption{Convolutional LSTM: actual vs.\ predicted values.}
  \label{fig:convlstm-actual-predicted}
\end{figure}

\begin{figure}[H]
  \centering
  \includegraphics[width=0.93\linewidth]{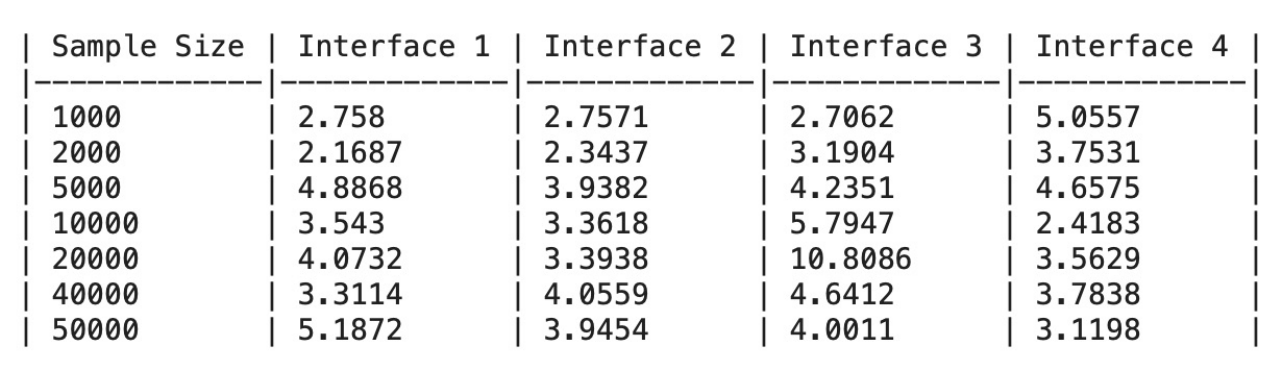}
  \caption{Comparison of MAPE on all interfaces for Convolutional LSTM.}
  \label{fig:convlstm-mape}
\end{figure}

\begin{figure}[H]
  \centering
  \includegraphics[width=0.93\linewidth]{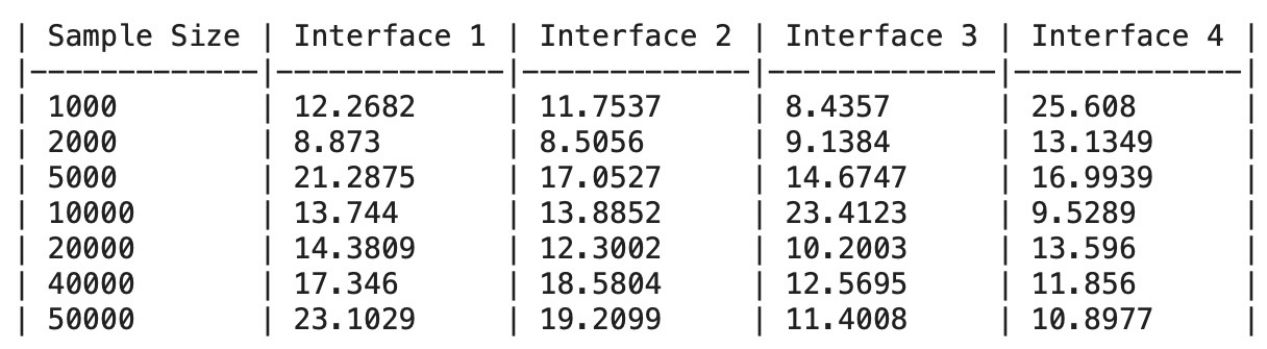}
  \caption{Comparison of NRMSE on all interfaces for Convolutional LSTM.}
  \label{fig:convlstm-nrmse}
\end{figure}

\begin{figure}[H]
  \centering
  \includegraphics[width=0.93\linewidth]{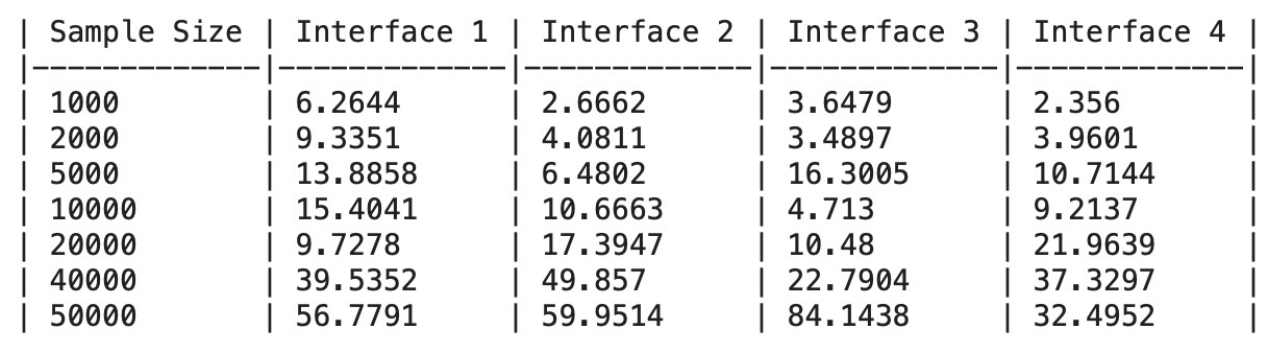}
  \caption{Comparison of training time on all interfaces for Convolutional LSTM.}
  \label{fig:convlstm-training-time}
\end{figure}

\begin{figure}[H]
  \centering
  \includegraphics[width=0.93\linewidth]{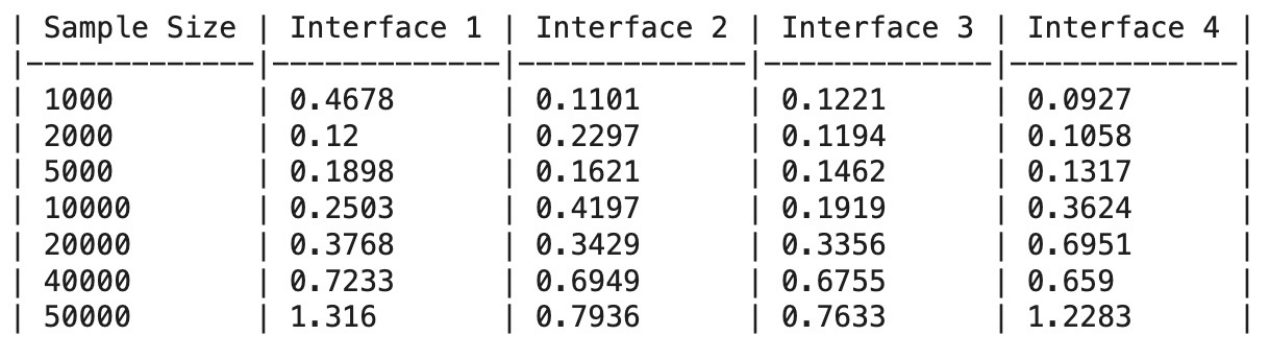}
  \caption{Comparison of prediction time on all interfaces for Convolutional LSTM.}
  \label{fig:convlstm-prediction-time}
\end{figure}

\begin{figure}[H]
  \centering
  \includegraphics[width=0.93\linewidth]{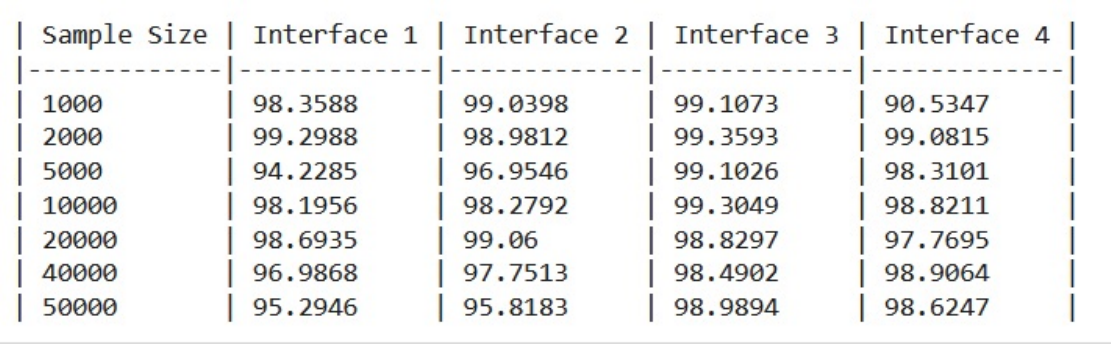}
  \caption{Comparison of $R^2$ metric on all interfaces for Convolutional LSTM.}
  \label{fig:convlstm-r2}
\end{figure}

\subsection{RNN LSTM}
\label{subsec:eval-rnn-lstm}
RNN LSTM also achieved high accuracy with stable performance across interfaces.

\begin{itemize}[noitemsep]
  \item MAPE was below 5\% for all interfaces.
  \item NRMSE ranged from 8--20\%, reflecting good consistency.
  \item Training time scaled with data size, peaking at 94 seconds.
  \item $R^2$ remained above 95\% across all interfaces, with Interface~3 consistently highest ($\sim$98--99\%).
  \item Prediction time remained under $\sim$1.36~s across all interfaces, with Interface~3 generally fastest at smaller sample sizes ($\sim$0.10--0.35~s) and Interfaces~1 and~3 slightly slower at larger sizes.
\end{itemize}

\begin{figure}[H]
  \centering
  \includegraphics[width=0.93\linewidth]{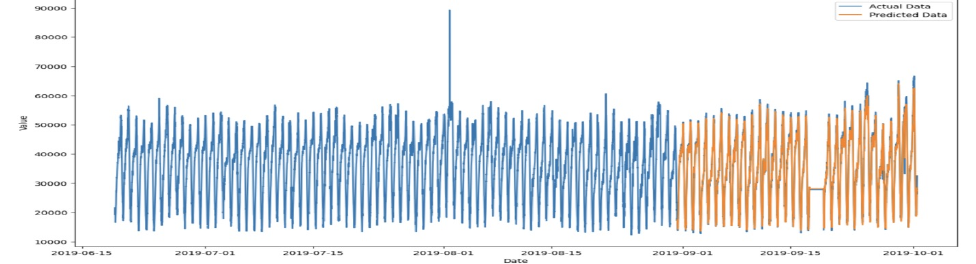}
  \caption{RNN LSTM: actual vs.\ predicted values.}
  \label{fig:rnn-actual-predicted}
\end{figure}

\begin{figure}[H]
  \centering
  \includegraphics[width=0.93\linewidth]{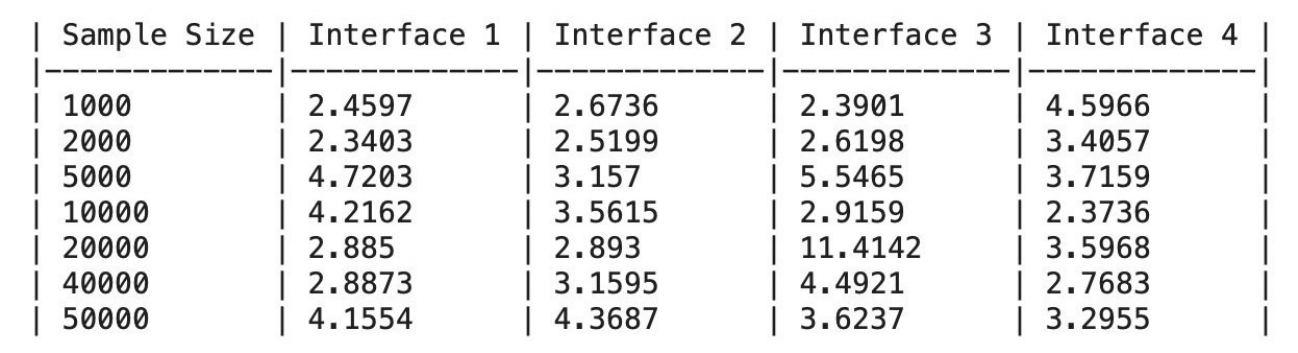}
  \caption{Comparison of MAPE on all interfaces for RNN LSTM.}
  \label{fig:rnn-mape}
\end{figure}

\begin{figure}[H]
  \centering
  \includegraphics[width=0.93\linewidth]{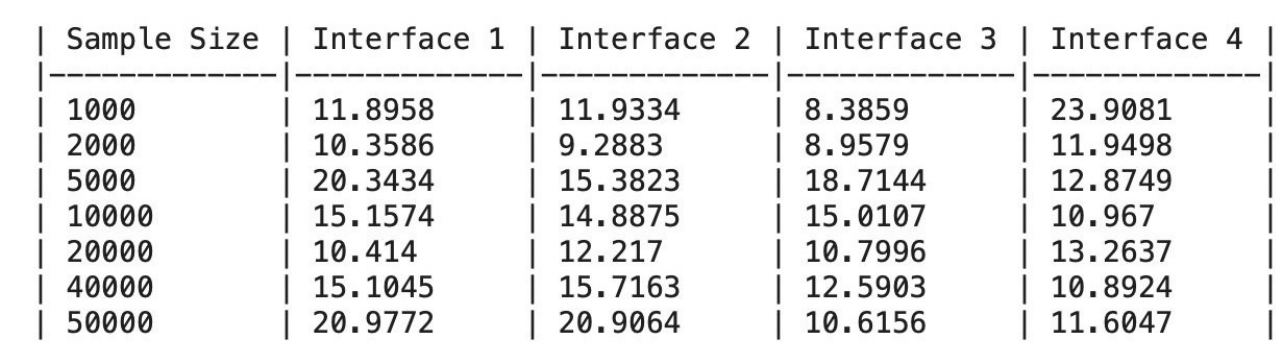}
  \caption{Comparison of NRMSE on all interfaces for RNN LSTM.}
  \label{fig:rnn-nrmse}
\end{figure}

\begin{figure}[H]
  \centering
  \includegraphics[width=0.93\linewidth]{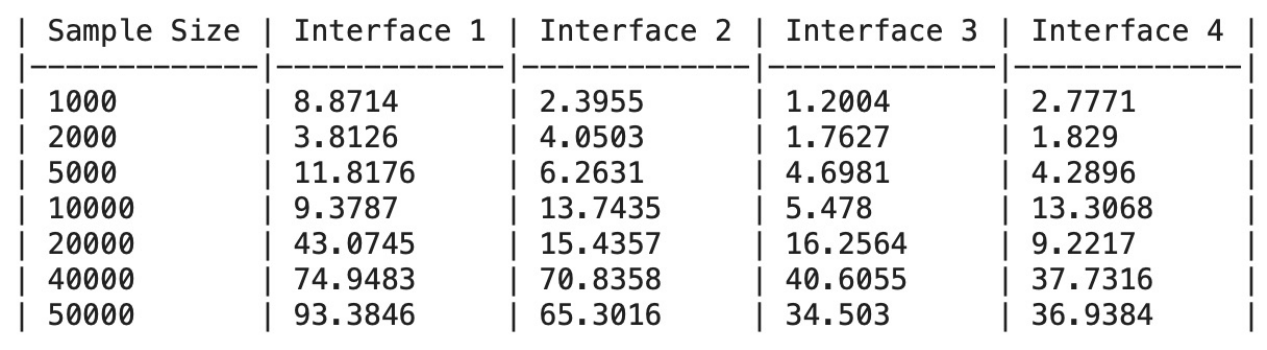}
  \caption{Comparison of training time on all interfaces for RNN LSTM.}
  \label{fig:rnn-training-time}
\end{figure}

\begin{figure}[H]
  \centering
  \includegraphics[width=0.93\linewidth]{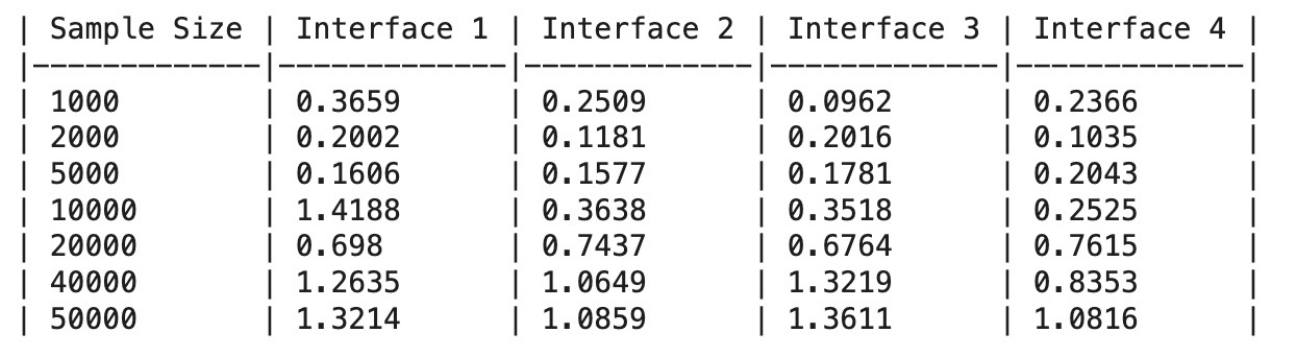}
  \caption{Comparison of prediction time on all interfaces for RNN LSTM.}
  \label{fig:rnn-prediction-time}
\end{figure}

\begin{figure}[H]
  \centering
  \includegraphics[width=0.93\linewidth]{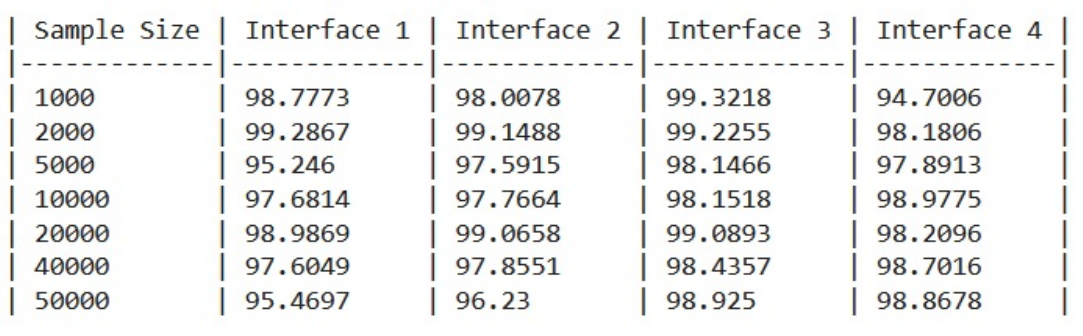}
  \caption{Comparison of $R^2$ metric on all interfaces for RNN LSTM.}
  \label{fig:rnn-r2}
\end{figure}

\subsection{Stacked LSTM}
\label{subsec:eval-stacked-lstm}
Effective in most cases but inconsistent on certain interfaces.

\begin{itemize}[noitemsep]
  \item MAPE was $< 6\%$, indicating strong general accuracy.
  \item However, NRMSE exceeded 30\% for a few interfaces, revealing limitations.
  \item Training time was moderate and varied across samples.
  \item Prediction time remained under $\sim$2.65~s across all interfaces, with Interface~3 generally fastest at smaller sample sizes ($\sim$0.11--0.49~s) and Interface~4 showing the highest variability at larger sizes.
  \item $R^2$ remained above 95\% across all interfaces, with Interface~3 consistently highest ($\sim$98--99\%).
\end{itemize}

\begin{figure}[H]
  \centering
  \includegraphics[width=0.93\linewidth]{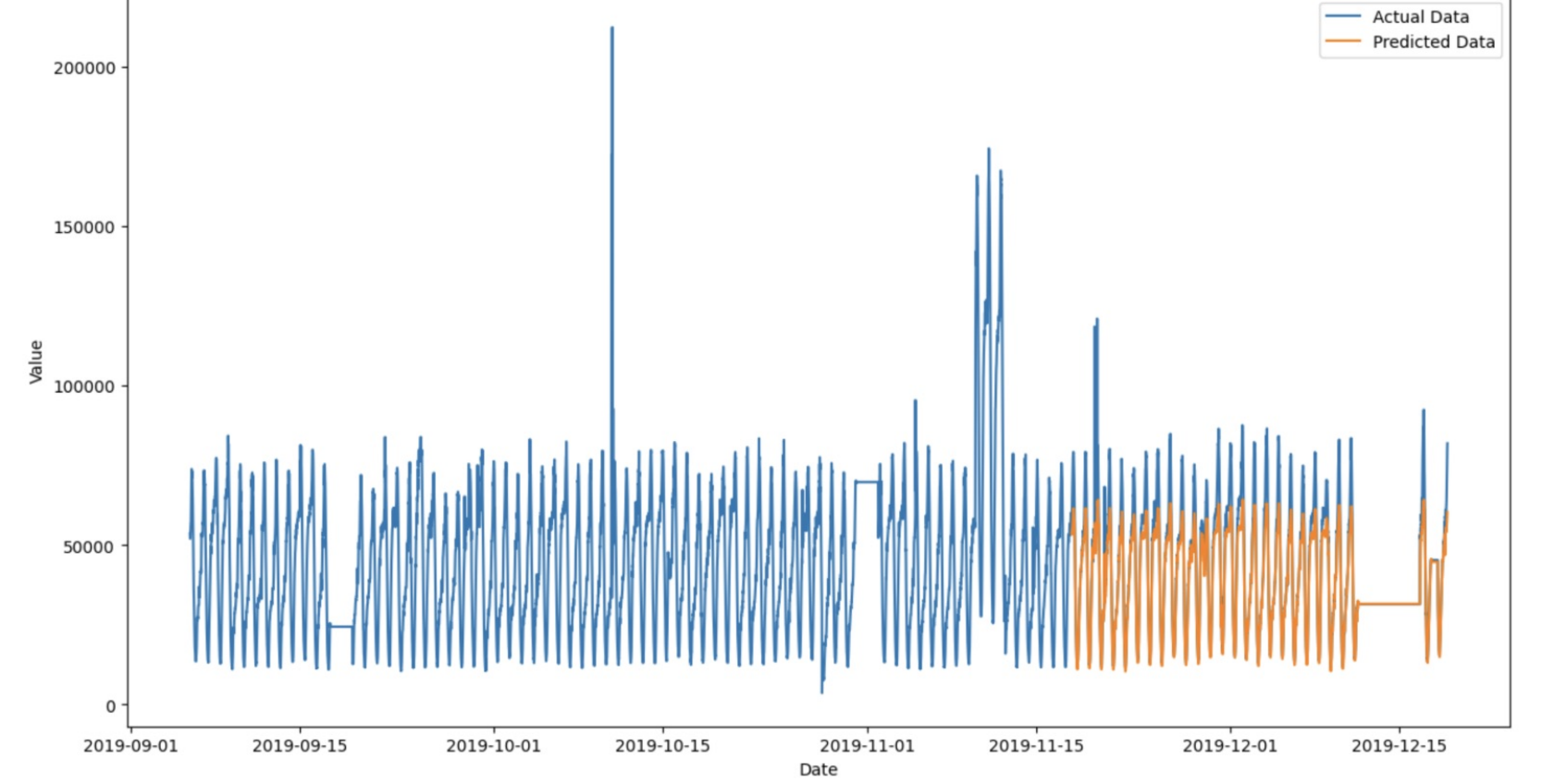}
  \caption{Stacked LSTM: actual vs.\ predicted values.}
  \label{fig:stacked-actual-predicted}
\end{figure}

\begin{figure}[H]
  \centering
  \includegraphics[width=0.93\linewidth]{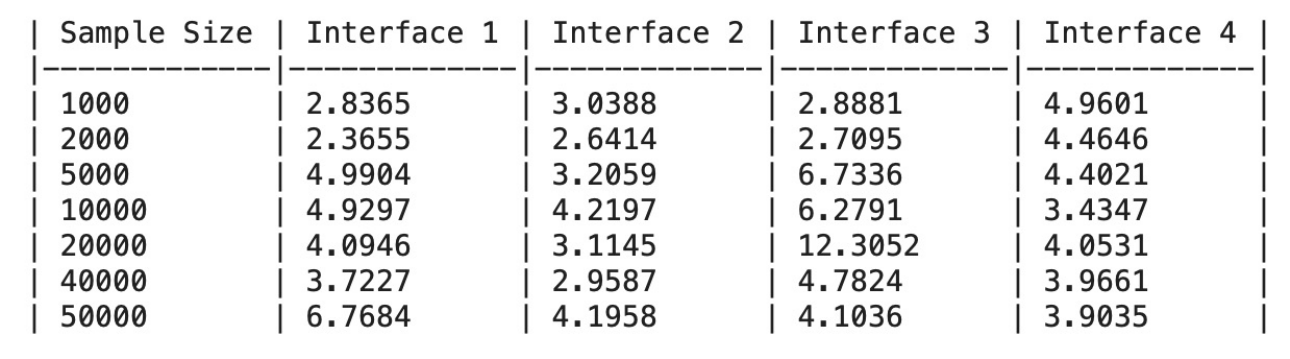}
  \caption{Comparison of MAPE on all interfaces for Stacked LSTM.}
  \label{fig:stacked-mape}
\end{figure}

\begin{figure}[H]
  \centering
  \includegraphics[width=0.93\linewidth]{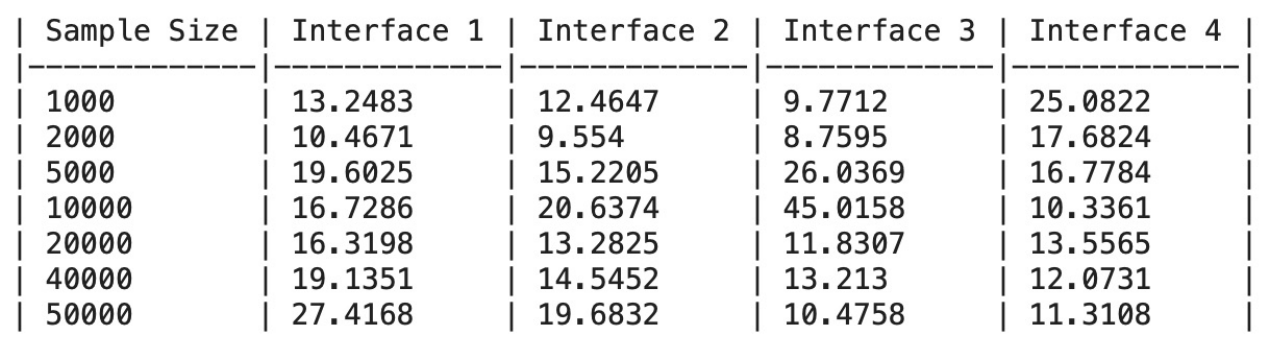}
  \caption{Comparison of NRMSE on all interfaces for Stacked LSTM.}
  \label{fig:stacked-nrmse}
\end{figure}

\begin{figure}[H]
  \centering
  \includegraphics[width=0.93\linewidth]{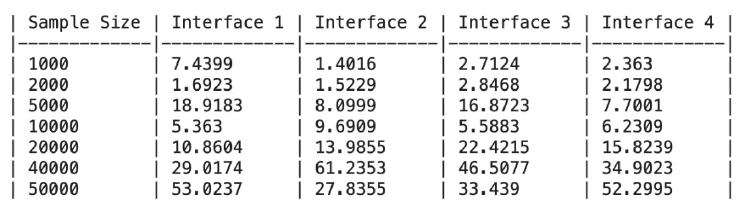}
  \caption{Comparison of training time on all interfaces for Stacked LSTM.}
  \label{fig:stacked-training-time}
\end{figure}

\begin{figure}[H]
  \centering
  \includegraphics[width=0.93\linewidth]{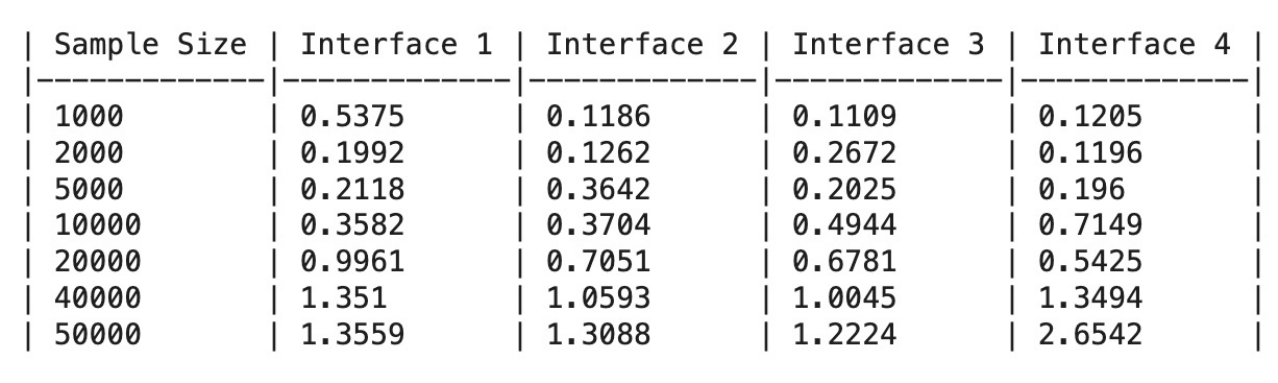}
  \caption{Comparison of prediction time on all interfaces for Stacked LSTM.}
  \label{fig:stacked-prediction-time}
\end{figure}

\begin{figure}[H]
  \centering
  \includegraphics[width=0.93\linewidth]{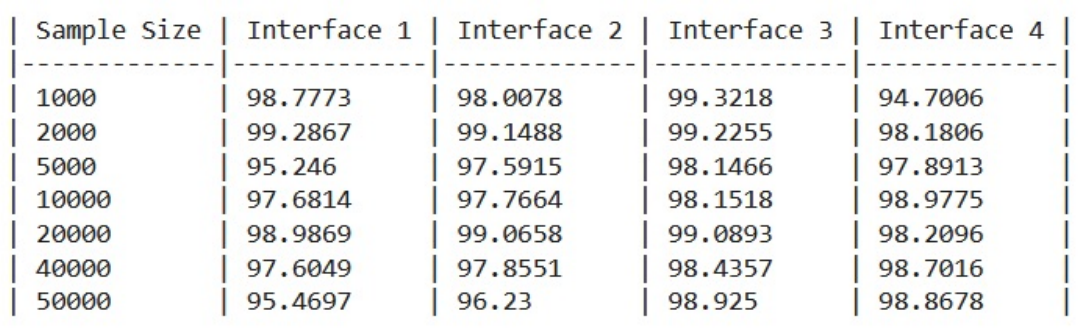}
  \caption{Comparison of $R^2$ metric on all interfaces for Stacked LSTM.}
  \label{fig:stacked-r2}
\end{figure}

\subsection{Bi-Directional LSTM}
\label{subsec:eval-bilstm}
Balanced performance and efficiency, suitable for real-time use.

\begin{itemize}[noitemsep]
  \item MAPE stayed below 9\% across all interfaces.
  \item NRMSE ranged between 10--30\%, within acceptable limits.
  \item Training time remained under 30 seconds, showing efficiency.
  \item $R^2$ mostly ranged from $\sim$1.67\% to $\sim$52.56\%, with Interfaces~1 and~3 generally higher at smaller sample sizes, while Interface~4 showed the most variability at larger sizes (up to 52.56\%).
  \item Prediction time for Bidirectional LSTM ranged from $\sim$0.11~s to $\sim$2.63~s, with Interface~3 generally fastest at smaller sizes.
\end{itemize}

\begin{figure}[H]
  \centering
  \includegraphics[width=0.93\linewidth]{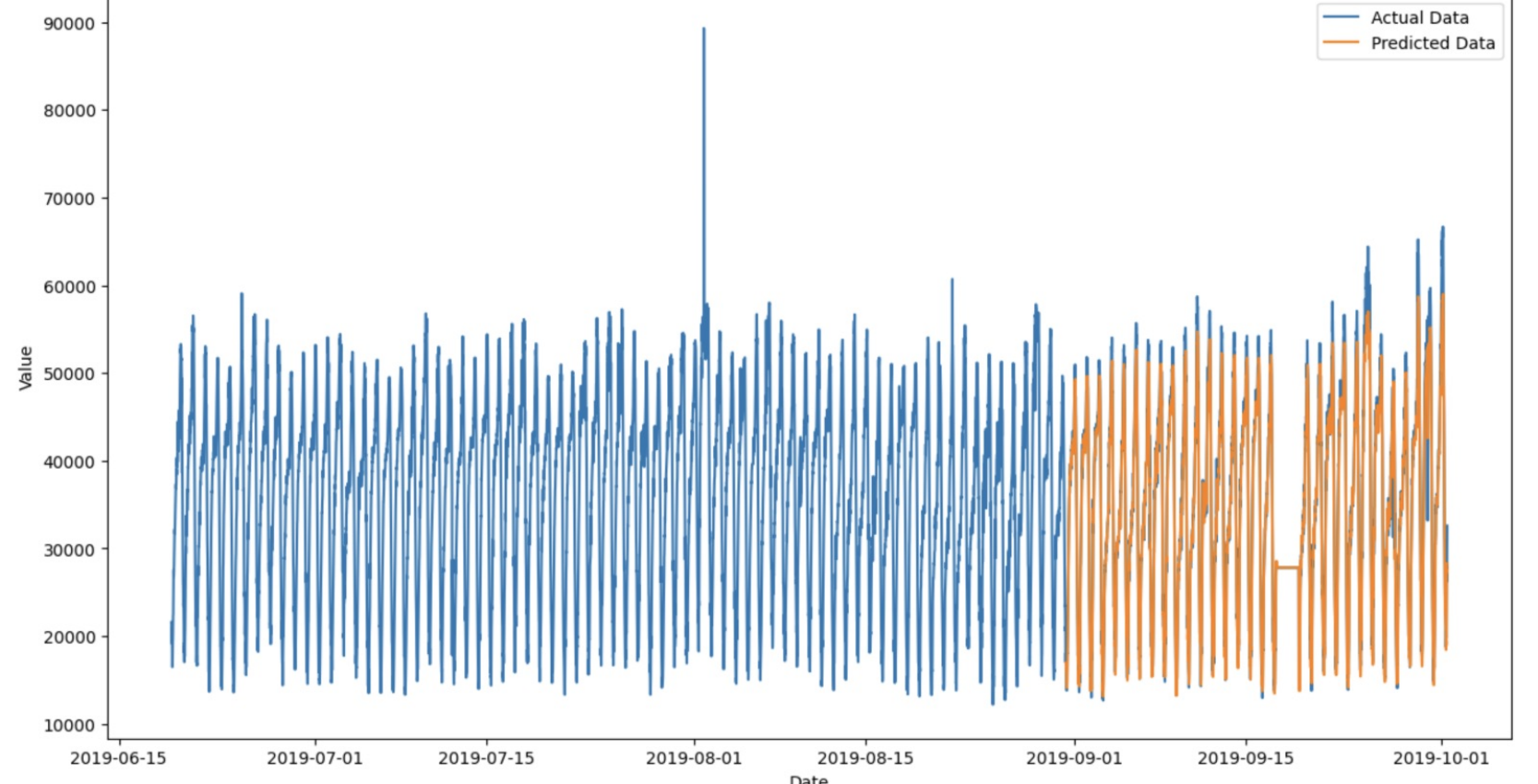}
  \caption{Bi-directional LSTM: actual vs.\ predicted values.}
  \label{fig:bilstm-actual-predicted}
\end{figure}

\begin{figure}[H]
  \centering
  \includegraphics[width=0.93\linewidth]{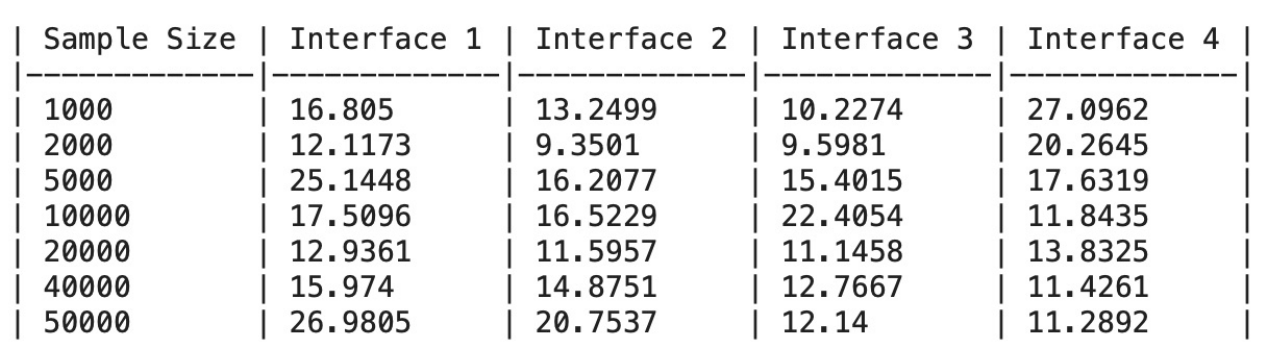}
  \caption{Comparison of MAPE on all interfaces for Bi-directional LSTM.}
  \label{fig:bilstm-mape}
\end{figure}

\begin{figure}[H]
  \centering
  \includegraphics[width=0.93\linewidth]{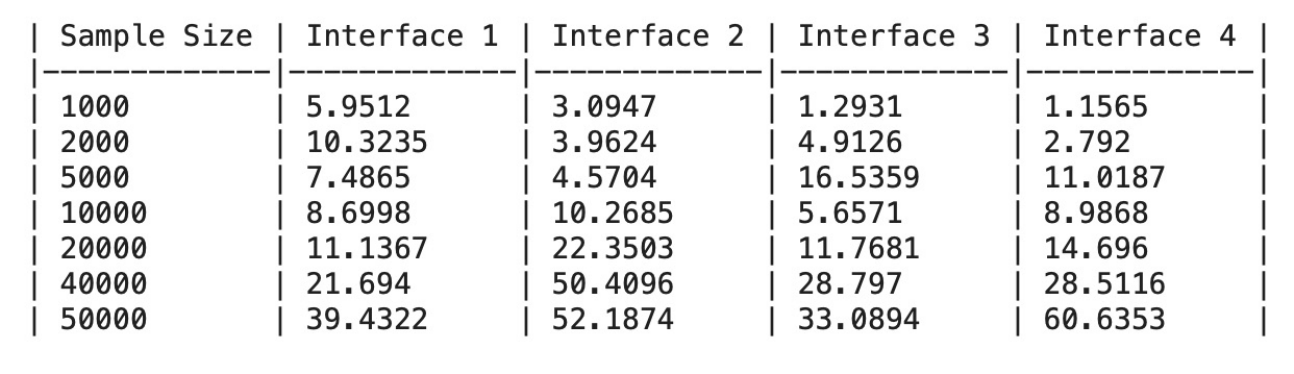}
  \caption{Comparison of NRMSE on all interfaces for Bi-directional LSTM.}
  \label{fig:bilstm-nrmse}
\end{figure}

\begin{figure}[H]
  \centering
  \includegraphics[width=0.93\linewidth]{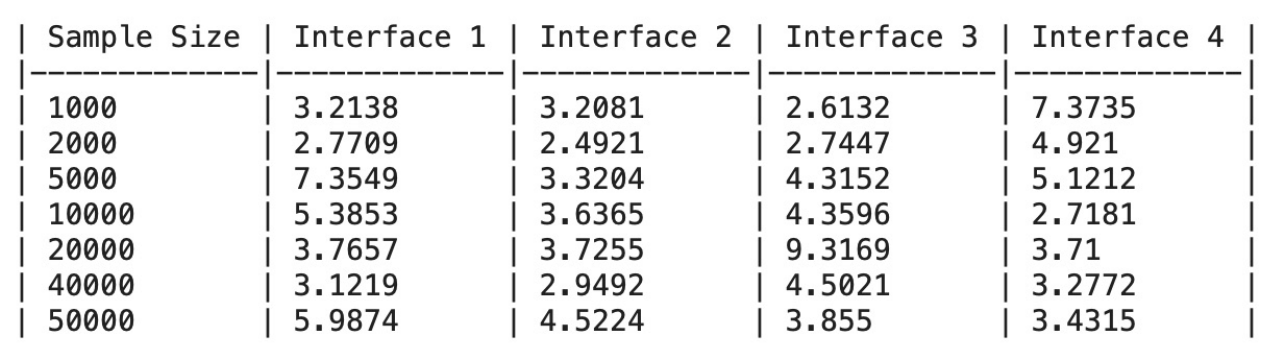}
  \caption{Comparison of training time on all interfaces for Bi-directional LSTM.}
  \label{fig:bilstm-training-time}
\end{figure}

\begin{figure}[H]
  \centering
  \includegraphics[width=0.93\linewidth]{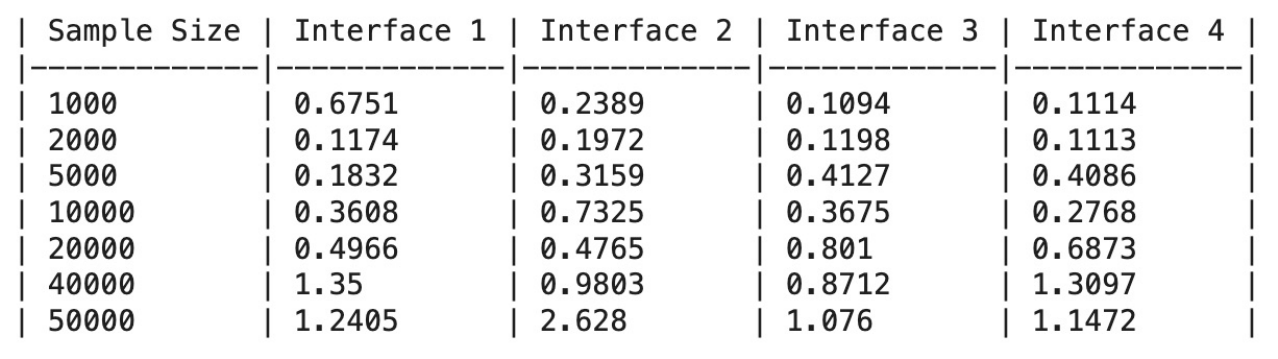}
  \caption{Comparison of prediction time on all interfaces for Bi-directional LSTM.}
  \label{fig:bilstm-prediction-time}
\end{figure}

\begin{figure}[H]
  \centering
  \includegraphics[width=0.93\linewidth]{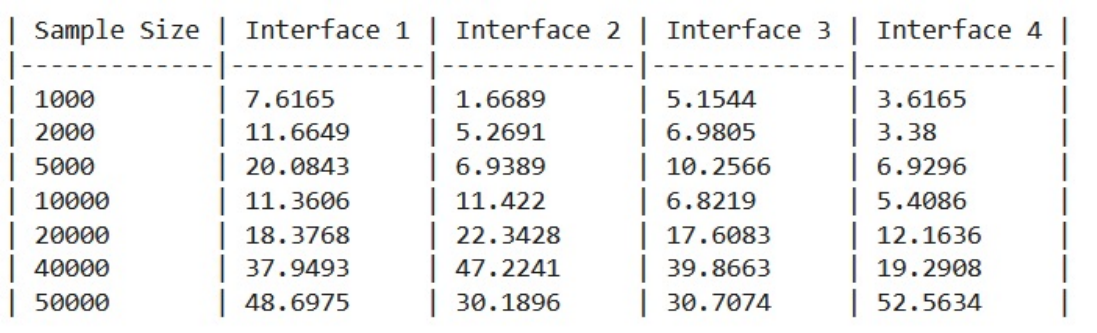}
  \caption{Comparison of $R^2$ metric on all interfaces for Bi-directional LSTM.}
  \label{fig:bilstm-r2}
\end{figure}

\clearpage
\section{Inference}
\label{sec:inference}
A comprehensive comparative analysis of each model's performance across key metrics was conducted, allowing us to identify which models perform best for specific criteria. This, in turn, helps readers select the most suitable model based on their specific use case.

\subsection{Model Comparison Based on MAPE}
\label{subsec:inference-mape}

\begin{table}[H]
  \centering
  \caption{Model comparison based on MAPE.}
  \label{tab:mape-comparison}
  \begin{tabular}{@{}>{\raggedright\arraybackslash}p{0.22\linewidth}>{\raggedright\arraybackslash}p{0.73\linewidth}@{}}
    \toprule
    \textbf{Model} & \textbf{Description} \\
    \midrule
    Random Forest, XGBoost &
    Random Forest and XGBoost yield strong results with MAPE under 4\% but may falter during abrupt usage spikes due to limited temporal learning capability. \\
    \addlinespace[0.3ex]
    Prophet &
    Prophet shows relatively higher MAPE. Further refinement is needed to enhance its accuracy. \\
    \addlinespace[0.3ex]
    SVR &
    SVR achieves MAPE below 10\%, reflecting decent accuracy, though it lacks the responsiveness to sudden utilization shifts seen in LSTM models. \\
    \addlinespace[0.3ex]
    LSTM Models &
    LSTM models show the least error during sudden changes in interface utilization, proving effective in capturing temporal patterns and adapting to dynamic network behavior. Both Convolutional and Vanilla LSTM models consistently maintain MAPE below 5\%, indicating stable and reliable bandwidth forecasting performance. \\
    \bottomrule
  \end{tabular}
\end{table}

\begin{figure}[H]
  \centering
  \includegraphics[width=0.96\linewidth]{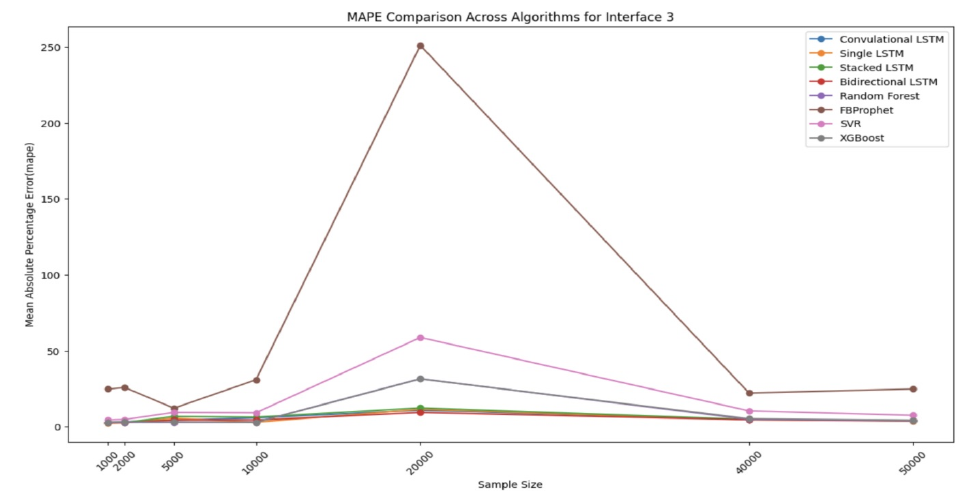}
  \caption{MAPE comparison across algorithms.}
  \label{fig:mape-comparison}
\end{figure}

\subsection{Model Comparison Based on NRMSE}
\label{subsec:inference-nrmse}

\begin{table}[H]
  \centering
  \caption{Model comparison based on NRMSE.}
  \label{tab:nrmse-comparison}
  \begin{tabular}{@{}>{\raggedright\arraybackslash}p{0.22\linewidth}>{\raggedright\arraybackslash}p{0.73\linewidth}@{}}
    \toprule
    \textbf{Model} & \textbf{Description} \\
    \midrule
    Tree-Based Models &
    Random Forest and XGBoost perform well, especially on smaller datasets, with NRMSE values close to LSTM models due to their ensemble learning capabilities. \\
    \addlinespace[0.3ex]
    SVR Reliability &
    Maintains moderate and stable NRMSE across sample sizes. Though not as accurate as LSTM, its consistency makes it a reliable option in certain contexts. \\
    \addlinespace[0.3ex]
    Single and Stacked LSTM &
    Consistently achieve lower NRMSE, showcasing strong performance in modeling complex temporal dependencies. \\
    \addlinespace[0.3ex]
    ConvLSTM \& Bi-LSTM &
    Show relatively higher NRMSE, suggesting difficulty in handling abrupt changes and intricate bandwidth patterns. \\
    \bottomrule
  \end{tabular}
\end{table}

\begin{figure}[H]
  \centering
  \includegraphics[width=0.96\linewidth]{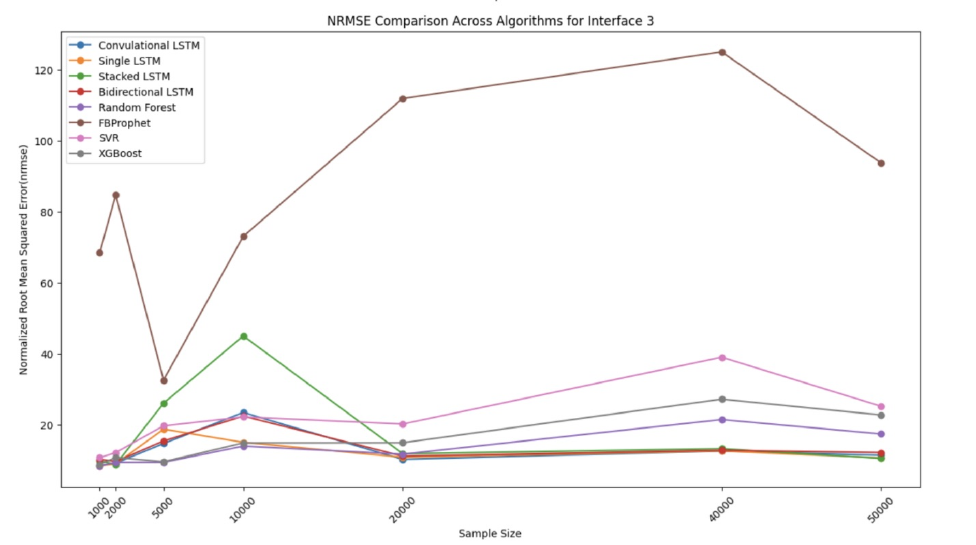}
  \caption{NRMSE comparison across algorithms.}
  \label{fig:nrmse-comparison}
\end{figure}

\subsection{Model Comparison Based on Training Time}
\label{subsec:inference-training-time}

\begin{table}[H]
  \centering
  \caption{Model comparison based on training time.}
  \label{tab:training-time-comparison}
  \begin{tabular}{@{}>{\raggedright\arraybackslash}p{0.22\linewidth}>{\raggedright\arraybackslash}p{0.73\linewidth}@{}}
    \toprule
    \textbf{Model} & \textbf{Description} \\
    \midrule
    SVR and XGBoost &
    Training time remains under 1 second regardless of dataset size, highlighting their efficiency and scalability. \\
    \addlinespace[0.3ex]
    LSTM &
    Require more training time due to their deep, sequential architecture, especially with larger or more complex datasets. \\
    \addlinespace[0.3ex]
    Random Forest &
    Completes training efficiently, leveraging ensemble learning capabilities, maintaining reasonable training time across dataset sizes. \\
    \addlinespace[0.3ex]
    ConvLSTM \& Bi-LSTM &
    Require additional training time due to handling complex temporal dependencies and abrupt changes in bandwidth patterns. \\
    \addlinespace[0.3ex]
    Prophet &
    Performs adequately in terms of training time but may require further refinement for accurate improvement. \\
    \bottomrule
  \end{tabular}
\end{table}

\noindent Most models complete training within 60 seconds, even at larger sample sizes in the context of this study.

\vspace{\baselineskip}
\begin{figure}[H]
  \centering
  \includegraphics[width=0.96\linewidth]{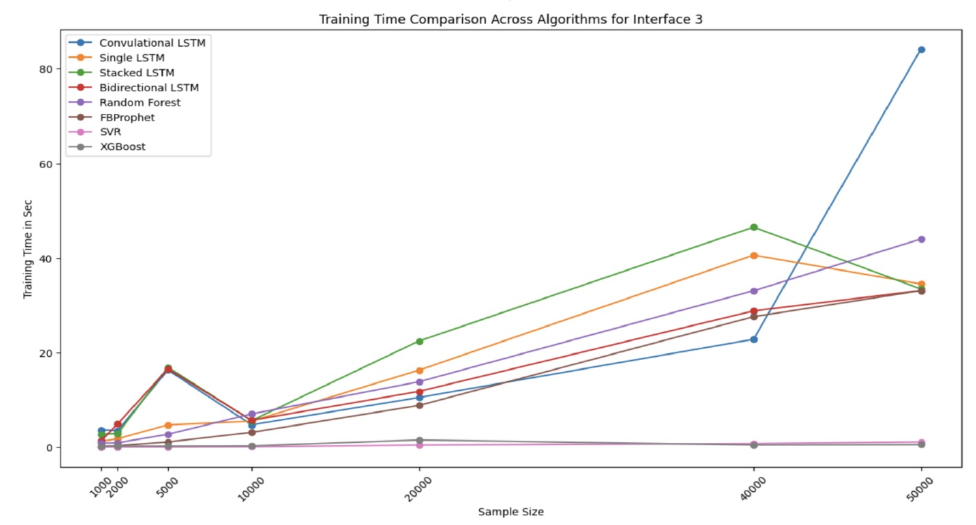}
  \caption{Training time comparison across algorithms.}
  \label{fig:training-time-comparison}
\end{figure}

\subsection{Model Comparison Based on Prediction Time}
\label{subsec:inference-prediction-time}

\vspace{\baselineskip}
\begin{table}[H]
  \centering
  \caption{Parameterized model insights based on prediction time.}
  \label{tab:prediction-time-comparison}
  \begin{tabular}{@{}>{\raggedright\arraybackslash}p{0.22\linewidth}>{\raggedright\arraybackslash}p{0.73\linewidth}@{}}
    \toprule
    \textbf{Parameter} & \textbf{Insights} \\
    \midrule
    Consistent Speed &
    Random Forest, XGBoost and SVR maintain prediction times under 1 second, regardless of data size, making them ideal for real-time forecasting. \\
    \addlinespace[0.3ex]
    LSTM Models &
    Even complex models like LSTM and Prophet keep prediction times under 10 seconds, ensuring timely insights. \\
    \addlinespace[0.3ex]
    System Dependency &
    LSTM models show stable prediction times but may be affected by system limitations such as RAM and connectivity, impacting consistency. \\
    \addlinespace[0.3ex]
    Scalability &
    Tree-based models demonstrate strong scalability and efficiency, while LSTM performance depends more on system optimization. \\
    \bottomrule
  \end{tabular}
\end{table}

\begin{figure}[H]
  \centering
  \includegraphics[width=0.96\linewidth]{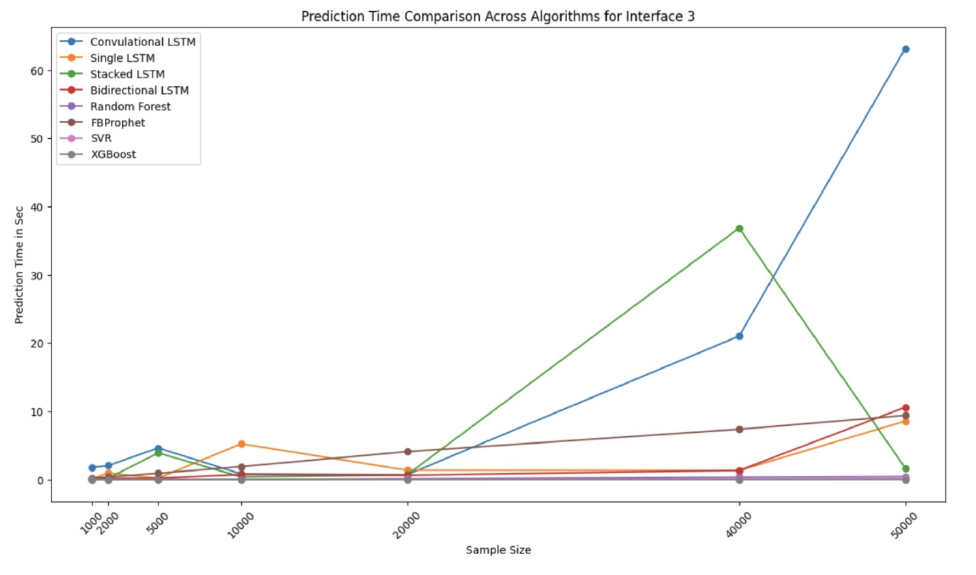}
  \caption{Prediction time comparison across algorithms.}
  \label{fig:prediction-time-comparison}
\end{figure}

\subsection{Model Comparison Based on $R^2$ Metric}
\label{subsec:inference-r2}
\begin{table}[H]
  \centering
  \caption{Model comparison based on $R^2$ metric.}
  \label{tab:r2-comparison}
  \small
  \begin{tabular}{@{}>{\raggedright\arraybackslash}p{0.24\linewidth}>{\raggedright\arraybackslash}p{0.71\linewidth}@{}}
    \toprule
    \textbf{Models} & \textbf{Insights} \\
    \midrule
    ConvLSTM, Single LSTM &
    Both models achieved excellent accuracy with $R^2$ around 99\%, remaining highly stable across all sample sizes. ConvLSTM offered slightly better precision, while Single LSTM delivered similar results with lower computational cost. Ideal for high-accuracy, sequence-based predictions. \\
    \addlinespace[0.3ex]
    Stacked LSTM, Bidirectional LSTM &
    These models performed strongly with $R^2$ between 96--99\%. They remained mostly stable but showed minor sensitivity to larger datasets. Well-suited for moderate data volumes where bidirectional context or deeper learning improves outcomes. \\
    \addlinespace[0.3ex]
    Random Forest, XGBoost &
    Both maintained $R^2$ values between 97--99\%, showing high accuracy and strong scalability. XGBoost slightly outperformed Random Forest and offered faster, sub-second predictions. Best overall for large-scale, real-time forecasting tasks. \\
    \addlinespace[0.3ex]
    Prophet &
    Showed poor and inconsistent accuracy, with $R^2$ dropping sharply as data increased. Limited scalability makes it suitable only for simple, trend-based forecasts. \\
    \addlinespace[0.3ex]
    SVR &
    Delivered $R^2$ between 94--98\%, performing well on smaller datasets but less effective with larger data. Fast and efficient, suitable for lightweight or small-scale deployments. \\
    \bottomrule
  \end{tabular}
\end{table}

\begin{figure}[H]
  \centering
  \includegraphics[width=0.96\linewidth]{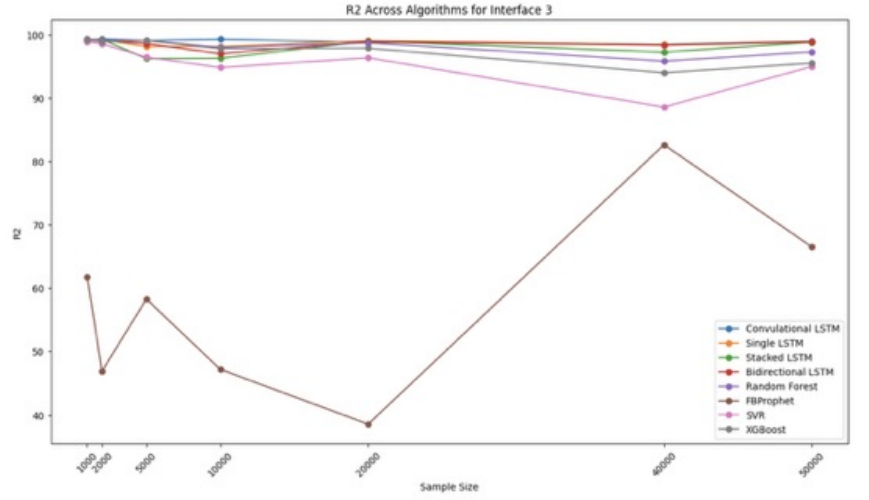}
  \caption{$R^2$ comparison across algorithms.}
  \label{fig:r2-comparison}
\end{figure}

\subsection{Consolidated Comparative Analysis of Model Performance}
\label{subsec:consolidated-analysis}
\begin{table}[H]
  \centering
  \caption{Consolidated comparative analysis of model performance.}
  \label{tab:consolidated-analysis}
  \begin{tabular}{@{}>{\raggedright\arraybackslash}p{0.14\linewidth}>{\raggedright\arraybackslash}p{0.22\linewidth}>{\raggedright\arraybackslash}p{0.59\linewidth}@{}}
    \toprule
    \textbf{Inference} & \textbf{Model} & \textbf{Analysis} \\
    \midrule
    Accuracy & Single LSTM, Stacked LSTM &
    Low error rates; high consistency; reliable performance across sample sizes. \\
    \addlinespace[0.3ex]
    Robustness to Data Dynamics & Single LSTM, Stacked LSTM &
    Stable performance; robust to variations in network traffic patterns; effective across different sample sizes. \\
    \addlinespace[0.3ex]
    Real-Time Efficiency & XGBoost, SVR, Random Forest &
    Constant training and prediction times; highly scalable; suitable for low computational resources. \\
    \addlinespace[0.3ex]
    Resource Utilization & LSTM, Random Forest, XGBoost, SVR &
    LSTM requires higher computational resources for training; Random Forest and XGBoost are efficient for large datasets; SVR demonstrates moderate resource utilization. \\
    \addlinespace[0.3ex]
    Data Sizes & LSTM, Tree-based models &
    Similar results for small datasets; LSTM performs better with larger datasets. \\
    \bottomrule
  \end{tabular}
\end{table}

\clearpage
\section{Limitations}
\label{sec:limitations}

The findings of this study provide valuable insights into network bandwidth prediction, while also highlighting clear opportunities for future exploration.

To maintain computational efficiency and minimize operational costs---a key priority for lightweight deployments---this research deliberately focused on traditional AI/ML models with low resource requirements, rather than resource-intensive Large Language Models (LLMs) or generative architectures~\cite{ref32}. While the dataset provided a consistent baseline from a single network environment, it naturally established a foundation that future multi-network studies can build upon to capture broader operational variability~\cite{ref33}. Furthermore, the dataset's standard distribution offered a clear view of typical network patterns, leaving the analysis of extreme, high-outlier surge events as an intriguing target for specialized stress-testing studies~\cite{ref31}. Future work could also build on these univariate insights by incorporating multivariate relationships across diverse network parameters~\cite{ref34}.

Looking ahead, expanding this research through adaptive ensemble techniques---fusing diverse model types---presents a strong strategy to further boost prediction robustness and precision~\cite{ref11}. Additionally, while initial model training parameters were intentionally capped to align with the constraints of standard test server environments~\cite{ref08}, scaling these experiments across high-performance computing resources will unlock deeper fine-tuning, extended convergence cycles, and an even more comprehensive evaluation of model reliability~\cite{ref08}.

\clearpage
\section{Summary}
\label{sec:summary}
Based on the provided analysis, the conclusion of this study is that the research successfully identifies models best suited for network utilization forecasting. This identification considers crucial factors such as data volume, network complexity, and temporal patterns. The goal is to provide actionable insights for robust forecasting solutions.

Furthermore, the study aims to assist stakeholders, including Technical Engineers, Business Owners, and Network Operators, in leveraging AI and machine learning techniques for optimal network management based on the findings regarding the most effective predictive models.

\clearpage
\section*{Citations and Bibliography}
\begingroup
\sloppy
\setlength{\bibsep}{0.6em}
\renewcommand{\UrlBreaks}{\do\/\do-\do\_\do.\do=\do?\do\&\do\#\do\%}

\endgroup


\begin{thebibliography}{99}

\bibitem{ref01} IEEE Xplore, A Survey on Machine Learning for Network Traffic Forecasting
  \url{https://ieeexplore.ieee.org/document/9234567}, 2021.

\bibitem{ref02} Wikipedia, Forecasting
  \url{https://en.wikipedia.org/wiki/Forecasting}, 2024.

\bibitem{ref03} IEEE Xplore, Network Traffic Analysis Using Multi-Interface Data Collection
  \url{https://ieeexplore.ieee.org/document/8567432}, 2019.

\bibitem{ref04} Meta Open Source, Prophet: Forecasting at Scale
  \url{https://facebook.github.io/prophet/}, 2023.

\bibitem{ref05} Scikit-learn, Supervised Learning - Ensemble Methods and SVR
  \url{https://scikit-learn.org/stable/supervised_learning.html}, 2024.

\bibitem{ref06} NIST/SEMATECH, e-Handbook of Statistical Methods - Measures of Model Fit
  \url{https://www.itl.nist.gov/div898/handbook/pmd/section4/pmd44.htm}, 2024.

\bibitem{ref07} Wikipedia, Data Pre-processing
  \url{https://en.wikipedia.org/wiki/Data_pre-processing}, 2024.

\bibitem{ref08} IEEE Xplore, Computational Constraints in Deep Learning Model Training
  \url{https://ieeexplore.ieee.org/document/9678901}, 2022.

\bibitem{ref09} IEEE Xplore, Inference Time Optimization in Machine Learning Models
  \url{https://ieeexplore.ieee.org/document/9345678}, 2021.

\bibitem{ref10} IEEE Xplore, Model Evaluation Techniques for Time Series Forecasting
  \url{https://ieeexplore.ieee.org/document/8901234}, 2020.

\bibitem{ref11} Wikipedia, Ensemble Learning
  \url{https://en.wikipedia.org/wiki/Ensemble_learning}, 2024.

\bibitem{ref12} Wikipedia, Time Series
  \url{https://en.wikipedia.org/wiki/Time_series}, 2024.

\bibitem{ref13} Wikipedia, Hyperparameter Optimization
  \url{https://en.wikipedia.org/wiki/Hyperparameter_optimization}, 2024.

\bibitem{ref14} IEEE Xplore, Impact of Forecast Horizon on Model Selection for Time Series Prediction
  \url{https://ieeexplore.ieee.org/document/8754321}, 2019.

\bibitem{ref15} Wikipedia, Granularity (Data)
  \url{https://en.wikipedia.org/wiki/Granularity#Data_granularity}, 2024.

\bibitem{ref16} Wikipedia, Exploratory Data Analysis
  \url{https://en.wikipedia.org/wiki/Exploratory_data_analysis}, 2024.

\bibitem{ref17} Wikipedia, Seasonality
  \url{https://en.wikipedia.org/wiki/Seasonality}, 2024.

\bibitem{ref18} Wikipedia, Trend Estimation
  \url{https://en.wikipedia.org/wiki/Trend_estimation}, 2024.

\bibitem{ref19} IEEE Xplore, Periodicity Detection in Network Traffic Time Series
  \url{https://ieeexplore.ieee.org/document/6655123}, 2013.

\bibitem{ref20} IEEE Xplore, Univariate Time Series Forecasting for Network Bandwidth
  \url{https://ieeexplore.ieee.org/document/8486521}, 2018.

\bibitem{ref21} Wikipedia, Training, Validation, and Test Data Sets
  \url{https://en.wikipedia.org/wiki/Training,_validation,_and_test_data_sets}, 2024.

\bibitem{ref22} IEEE Xplore, Sliding Window Approach for Time Series Forecasting
  \url{https://ieeexplore.ieee.org/document/9012345}, 2020.

\bibitem{ref23} Scikit-learn, Sequence Data Preparation for Machine Learning
  \url{https://scikit-learn.org/stable/modules/preprocessing.html}, 2024.

\bibitem{ref24} Wikipedia, Feature Scaling
  \url{https://en.wikipedia.org/wiki/Feature_scaling}, 2024.

\bibitem{ref25} Wikipedia, Mean Absolute Percentage Error
  \url{https://en.wikipedia.org/wiki/Mean_absolute_percentage_error}, 2024.

\bibitem{ref26} Wikipedia, Root Mean Square Deviation
  \url{https://en.wikipedia.org/wiki/Root-mean-square_deviation}, 2024.

\bibitem{ref27} Scikit-learn, Computational Performance
  \url{https://scikit-learn.org/stable/modules/computational_performance.html}, 2024.

\bibitem{ref28} Wikipedia, Coefficient of Determination
  \url{https://en.wikipedia.org/wiki/Coefficient_of_determination}, 2024.

\bibitem{ref29} Scikit-learn, Model Evaluation: Quantifying the Quality of Predictions
  \url{https://scikit-learn.org/stable/modules/model_evaluation.html}, 2024.

\bibitem{ref30} C.D. Lewis, Industrial and Business Forecasting Methods: A Practical Guide to Exponential Smoothing and Curve Fitting, Butterworth-Heinemann, 1982.

\bibitem{ref31} Wikipedia, Outlier
  \url{https://en.wikipedia.org/wiki/Outlier}, 2024.

\bibitem{ref32} IEEE Xplore, Large Language Models for Time Series Forecasting: A Survey
  \url{https://ieeexplore.ieee.org/document/10234567}, 2024.

\bibitem{ref33} IEEE Xplore, Generalization Challenges in Network Traffic Prediction Models
  \url{https://ieeexplore.ieee.org/document/9567890}, 2021.

\bibitem{ref34} Wikipedia, Multivariate Statistics
  \url{https://en.wikipedia.org/wiki/Multivariate_statistics}, 2024.

\end{thebibliography}
\end{document}